\documentclass[10pt,twocolumn,letterpaper]{article}

\usepackage[pagenumbers]{cvpr} % To force page numbers, e.g. for an arXiv version

\definecolor{cvprblue}{rgb}{0.21,0.49,0.74}
\usepackage[pagebackref,breaklinks,colorlinks,allcolors=cvprblue]{hyperref}

\usepackage{tabularx} % 引入tabularx宏包
\usepackage{multirow} % 确保multirow宏包也被引入
\usepackage{booktabs}
\usepackage{amsmath}  
\usepackage{amssymb}  
\usepackage{mathrsfs} 
\usepackage{amsthm,amsfonts}
\usepackage{array}
\usepackage{float}
\usepackage{pifont} % \ding{xx}
\usepackage{bbding} % \Checkmark,\XSolid,...
\usepackage{fontawesome} % \faCheck,\faTime
\def\ie{\emph{i.e.}}
\def\eg{\emph{e.g.}}

\usepackage[dvipsnames]{xcolor}
\newcommand{\mname}{\textsl{ConSeg}}
\newcommand{\fname}{\textsl{CaM}}
\definecolor{ourpurple}{HTML}{7030A0} 
\definecolor{ouryellow}{HTML}{F4B402} 
\newcommand{\highlight}[1]{\textbf{\color{ourpurple}#1}}
\def\paperID{786} % *** Enter the Paper ID here
\def\confName{CVPR}
\def\confYear{2025}

\title{Code-as-Monitor: Constraint-aware Visual Programming for \\ Reactive and Proactive Robotic Failure Detection}

\author{
Enshen Zhou\textsuperscript{1}\footnotemark[1]~~\footnotemark[3]~, \quad
Qi Su\textsuperscript{2,3,4}\footnotemark[1]~, \quad
Cheng Chi\textsuperscript{3}\footnotemark[1]~~\footnotemark[2]~,\\
Zhizheng Zhang\textsuperscript{4}, \quad 
Zhongyuan Wang\textsuperscript{3}, \quad
Tiejun Huang\textsuperscript{2,3}, \quad
Lu Sheng\textsuperscript{1}\footnotemark[2]~, \quad
He Wang\textsuperscript{2,3,4}\footnotemark[2]\\
\small $^{1}$School of Software, Beihang University~~
\small$^{2}$School of Computer Science, Peking University~~ \\
\small$^{3}$Beijing Academy of Artificial Intelligence~~
\small$^{4}$Galbot\\
\tt\footnotesize zhouenshen@buaa.edu.cn~~~chicheng15@mails.ucas.ac.cn~~~lsheng@buaa.edu.cn~~~hewang@pku.edu.cn
}

\begin{document}
% \maketitle

\twocolumn[{%
\renewcommand\twocolumn[1][]{#1}%
\maketitle
% \vspace{-12mm}
\vspace{-10mm}
\begin{center}
    % \centering
    \captionsetup{type=figure}
    \includegraphics[width=1\linewidth]{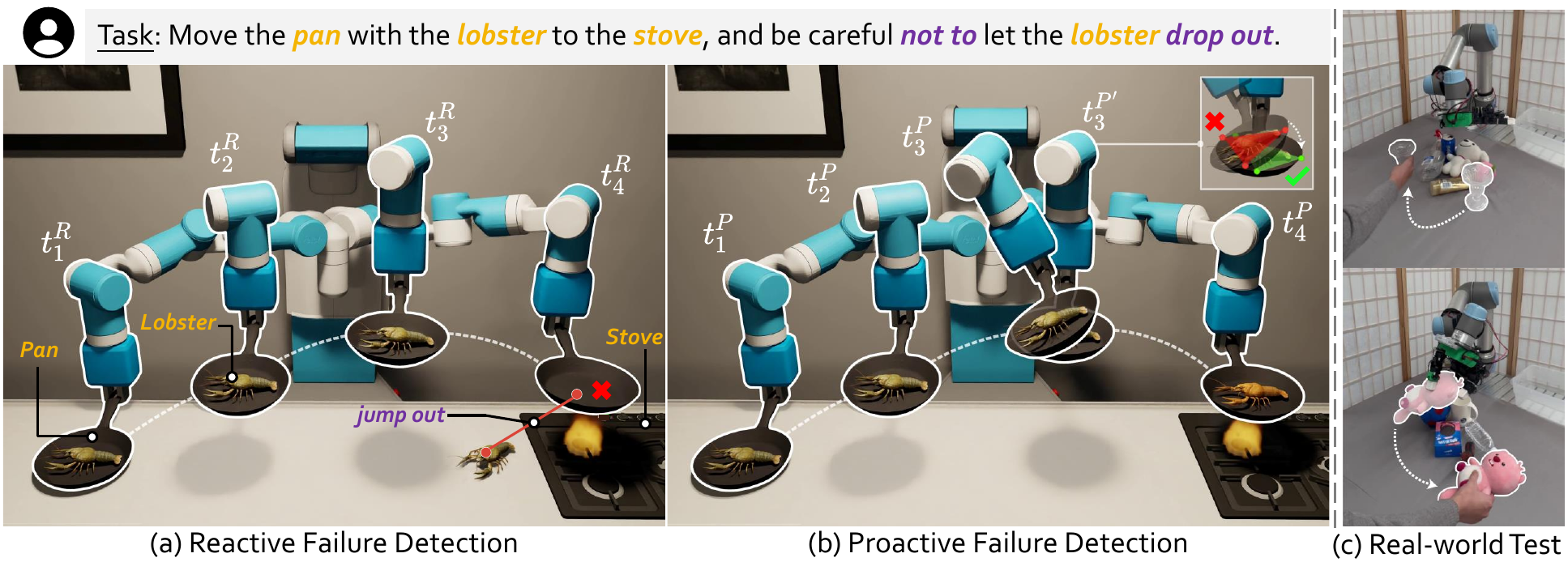}
    \vspace{-4mm}
    \captionof{figure}{    
    For the task ``\textit{Move the pan with lobster to the stove without losing the lobster}'', \textbf{(a)} reactive failure detection identifies failures after they occur, and \textbf{(b)} proactive failure detection prevents foreseeable failures. 
    In \textbf{(a)}, at \( t^R_4 \), the robot detects the failure after the lobster unpredictably jumps out due to the heat.
    In \textbf{(b)}, pan tilting is detected at \( t^P_3 \) and corrected it at \( t^{P'}_3 \), requiring real-time precision.
    We formulate both tasks as spatio-temporal constraint satisfaction problems, leveraging our proposed constraint elements for precise, real-time checking.
    For example, in \textbf{(a)}, a large relative distance between pan and lobster indicates failure; in \textbf{(b)}, a large angle between the pan and the horizontal plane needs correction.
    \textbf{(c)} shows that our method combined with an open-loop policy forms a closed-loop system, enabling proactive (\eg, detecting moving glass during grasping) and reactive (\eg, removing toy after grasping) failure detection in cluttered scenes.
    }\label{fig: motivation}
\end{center}%
}]

\let\thefootnote\relax\footnotetext{
$^*$ Equal contribution \hspace{5pt}$^\dagger$ Corresponding author}
\let\thefootnote\relax\footnotetext{ $^\ddagger$ Work done during internship at Galbot}

\begin{abstract}
% \vspace{-6mm}

Automatic detection and prevention of open-set failures are crucial in closed-loop robotic systems.
Recent studies often struggle to simultaneously identify unexpected failures reactively after they occur and prevent foreseeable ones proactively.
To this end, we propose Code-as-Monitor (CaM), a novel paradigm leveraging the vision-language model (VLM) for both open-set reactive and proactive failure detection. 
The core of our method is to formulate both tasks as a unified set of spatio-temporal constraint satisfaction problems and use VLM-generated code to evaluate them for real-time monitoring.
To enhance the accuracy and efficiency of monitoring, we further introduce constraint elements that abstract constraint-related entities or their parts into compact geometric elements.
This approach offers greater generality, simplifies tracking, and facilitates constraint-aware visual programming by leveraging these elements as visual prompts.
Experiments show that CaM achieves a $28.7\%$ higher success rate and reduces execution time by $31.8\%$ under severe disturbances compared to baselines across three simulators and a real-world setting.
Moreover, CaM can be integrated with open-loop control policies to form closed-loop systems, enabling long-horizon tasks in cluttered scenes with dynamic environments.
See the project page at \href{https://zhoues.github.io/Code-as-Monitor/}{https://zhoues.github.io/Code-as-Monitor}.

% Automatic detection and prevention of open-set failures are crucial in closed-loop robotic systems. Recent studies often struggle to simultaneously identify unexpected failures reactively after they occur and prevent foreseeable ones proactively. To this end, we propose Code-as-Monitor (CaM), a novel paradigm leveraging the vision-language model (VLM) for both open-set reactive and proactive failure detection. The core of our method is to formulate both tasks as a unified set of spatio-temporal constraint satisfaction problems and use VLM-generated code to evaluate them for real-time monitoring. To enhance the accuracy and efficiency of monitoring, we further introduce constraint elements that abstract constraint-related entities or their parts into compact geometric elements. This approach offers greater generality, simplifies tracking, and facilitates constraint-aware visual programming by leveraging these elements as visual prompts. Experiments show that CaM achieves a 28.7% higher success rate and reduces execution time by 31.8% under severe disturbances compared to baselines across three simulators and a real-world setting. Moreover, CaM can be integrated with open-loop control policies to form closed-loop systems, enabling long-horizon tasks in cluttered scenes with dynamic environments.

%
% Robotic Failure Detection, Vision-Language Model, Visual Programming, Visual Prompting
%
\end{abstract} 
% \vspace{-4mm}
\section{Introduction}
\label{sec: intro}
% \clearpage
%
% Automatically preventing and detecting execution failures is crucial for closed-loop robotic systems performing long-horizon, open-world tasks.
%
As expectations grow for robots to handle long-horizon tasks within intricate environments, failures are unavoidable. 
%
% Therefore, it is vital for robotic systems to detect and prevent those failures automatically in a closed-loop manner. 
%
Therefore, automatically detecting and preventing those failures plays a vital role in ensuring the tasks can eventually be solved, especially for closed-loop robotic systems.
There are two modes of failure detection~\cite{lemasurier2024reactive}, \textit{reactive} and \textit{proactive}. 
As depicted in \cref{fig: motivation}, \textit{reactive} failure detection aims to identify failures after they occur~(\eg, recognizing that lobster lands on the table, \textit{indicating} a delivery failure).
In contrast, \textit{proactive} failure detection aims to prevent foreseeable failures~(\eg, recognizing that a tilted pan could cause the lobster to fall out, \textit{leading to} a delivery failure).
Both detection modes are even more challenging in open-set scenarios, where the failures are not predefined.
%

% Recent studies primarily focus on \textit{reactive} failure detection~\cite{huang2022inner, silver2024generalized, skreta2023errors, du2023vision, liu2024self, zheng2024evaluating, duan2024aha, dai2024racer, guo2023doremi, yu2023multireact}, 
% with the help of large language models~(LLMs)~\cite{dubey2024llama, touvron2023llama} or vision-language models~(VLMs)~\cite{achiam2023gpt, liu2024visual} to identify failures after they occur, in the form of visual question-answering~(VQA). 
%
% These methods relax real-time performance and precision requirements, aligning with LLMs/VLMs' inherent limitations such as high computational cost and inadequate 3D spatio-temporal understanding.
With the help of large language models~(LLMs)~\cite{dubey2024llama, touvron2023llama} and vision-language models~(VLMs)~\cite{achiam2023gpt, liu2024visual}, recent studies can achieve open-set \textit{reactive} failure detection~\cite{huang2022inner, silver2024generalized, skreta2023errors, du2023vision, liu2024self, zheng2024evaluating, duan2024aha, dai2024racer, guo2023doremi, yu2023multireact} as a special case of visual question answering~(VQA) tasks. 
However, these methods often bear compromised execution speeds and coarse-grained detection accuracy, due to the high computational costs and inadequate 3D spatio-temporal perception capability of recent LLMs/VLMs.
%
% But recent LLMs/VLMs' limitations such as high computational costs and inadequate 3D spatio-temporal perception capability, 
%
Moreover, open-set \textit{proactive} failure detection, which has been less explored in the literature, presents even severer challenges as it is required to foresee potential causes of failure and monitor them in real-time with high precision to anticipate and prevent imminent failure.
Simply adapting LLMs/VLMs cannot meet such expectations.

%
% In this work, we aim to study how to use a generalized modeling approach to achieve both challenging proactive and reactive failure detection. 
%
In this work, we aim to develop an open-set failure detection framework that achieves reactive and proactive detection simultaneously, not only benefiting from the generalization power of VLMs but also enjoying high precision in monitoring failure characteristics with real-time efficiency. 
We address this by formulating both tasks as a unified set of spatio-temporal constraint satisfaction problems, which can be \emph{precisely} translated by VLMs into executable programs.
Such visual programs can efficiently verify whether entities (\eg, robots, objects) or their parts in the captured environment maintain or achieve required states during or after execution (\ie, satisfying constraints), so as to \emph{immediately} prevent or detect failures.
To the best of our knowledge, this is the first attempt to integrate both detection modes within a single framework.
We name this constraint-aware visual programming framework as Code-as-Monitor~({\fname}).
%
% To enable real-time and precise constraint checking, we leverage VLMs for automatically generating code to compute inter-entity constraints and monitor their fulfillment, naming the framework Code-as-Monitor~({\fname}).

% To be specific, the proposed spatio-temporal constraint computation is designed to replace naively applying costly and unreliable VLMs.
%
To be specific, the proposed spatio-temporal constraint satisfaction scheme abstracts the constraint-related entity or part segments from the observed images into compact geometric elements (\eg, points, lines, and surfaces), as shown in \cref{fig: motivation}.
It simplifies the monitoring of constraint satisfaction by tracking and evaluating the spatio-temporal combinational dynamics of these elements, eliminating the most irrelevant geometric and visual details of the raw entities/parts.
The constraint element detection and tracking are grounded by our trained constraint-aware segmentation and off-the-shelf tracking models, ensuring speed, accuracy, and certain open-set adaptation capabilities.
The evaluation protocol is in the form of VLM-generated code, \ie, monitor code, which is visually prompted by the starting frames of a sub-goal and its associated constraint elements, in addition to the textual constraints that must be fulfilled. 
Once generated, this monitor code could detect reactive or proactive failures just by being executed according to the tracked constraint elements, without needing to call the VLMs again.
Therefore, this minimalist scheme is generalizable to open-set failure detection for unseen entities and scenes (enabled by the potential diversity of the structured associations of constraint elements) as well as common skills (powered by the rich prior knowledge offered by VLMs), maintaining sufficient detection accuracy and real-time execution speed. 

We conduct extensive experiments in three simulators~(\ie, CLIPort~\cite{shridhar2021cliport}
, Omnigibson~\cite{li2022behavior}, and RLBench~\cite{james2019rlbench}) and one real-world setting, spanning diverse manipulation tasks (\eg, pick \& place, articulated objects, tool-use), robot platforms (\eg, UR5, Fetch, Franka) and end-effectors (\eg, suction cup, gripper, dexterous hand).
The results show that {\fname} is generalizable and achieves both \textit{reactive} and \textit{proactive} failure detection in real-time, resulting in $28.7\%$ higher success rates and reduced execution time by $31.8\%$ under severe disturbances compared to the baselines.
Moreover, in \cref{subsec: main results in real world}, {\fname} can be integrated with the existing open-loop control policy to form a closed-loop system, enabling long-horizon tasks in cluttered scenes with environment dynamics and human disturbances.
Our contributions are summarized as follows:
\begin{itemize}
    \item We introduce Code-as-Monitor~(\fname), a novel paradigm that leverages VLMs for both \textit{reactive} and \textit{proactive} failure detection via constraint-aware visual programming.
    \item We propose the constraint elements to enhance the accuracy and efficiency of constraint satisfaction monitoring.
    \item Extensive experiments show that {\fname} is generalizable and achieves more real-time and precise failure detection than baselines across simulators and real-world settings. 
\end{itemize}
\section{Related work}
\label{sec: related work}

\begin{figure*}[t]
\centering
\includegraphics[width=\linewidth]{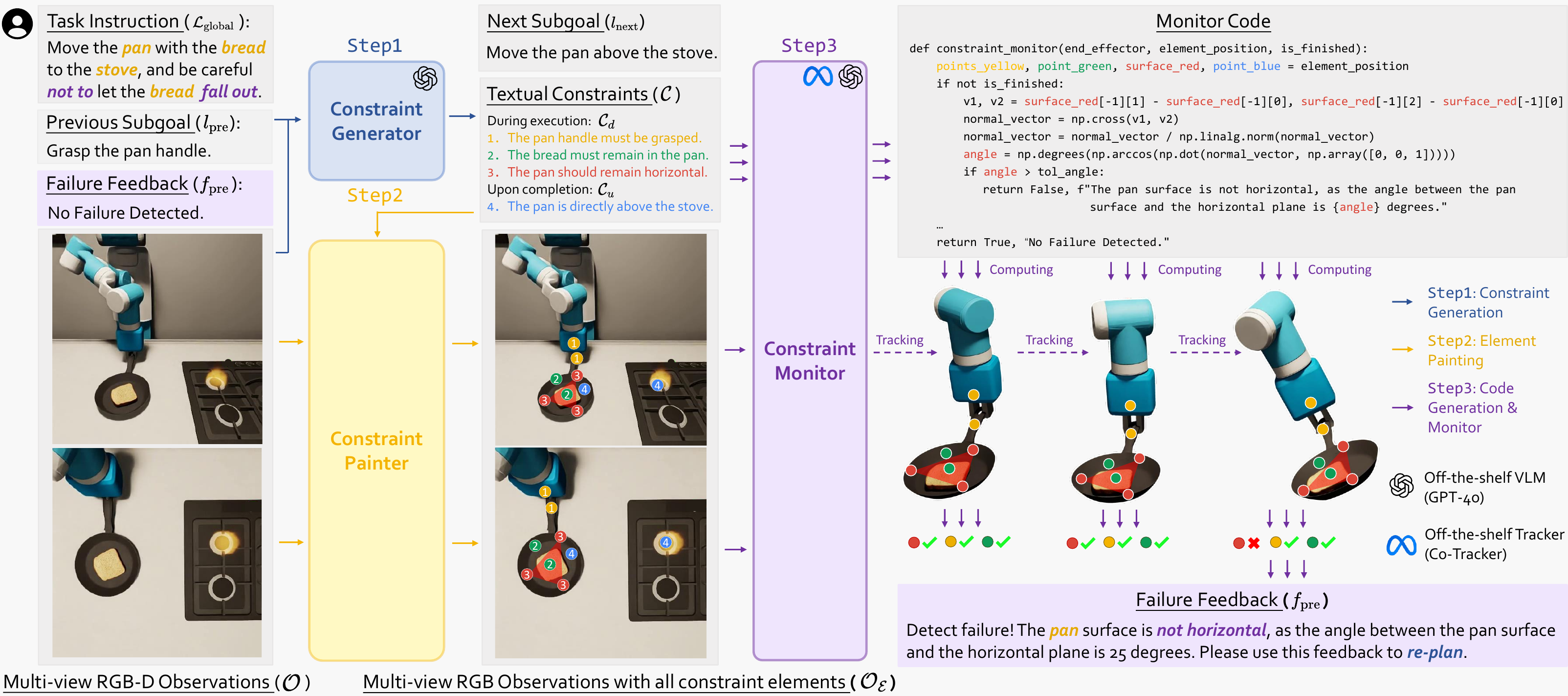}
\vspace{-6mm}
   \caption{
   Overview of Code-as-Monitor. Given task instructions and prior information, the Constraint Generator derives the next subgoal and corresponding textual constraints based on multi-view observations. 
   The Painter maps these constraints onto images as constraint elements. 
   The Monitor generates monitor code from these images and tracks them for real-time monitoring. 
   If any constraint is violated, it outputs the reason for failure and triggers re-planning. 
   This framework unifies \textit{reactive} and \textit{proactive} failure detection via constraints, more generally abstracts relevant entities/parts through constraint elements, and ensures precise and real-time monitoring via code evaluation.}
\label{fig: pipeline}
% \vspace{-5mm}
\vspace{-3mm}
\end{figure*}

\noindent\textbf{Robotic Failure Detection.}
Recent advances in LLMs~\cite{dubey2024llama, touvron2023llama, brown2020language, touvron2023llama2} and VLMs~\cite{achiam2023gpt, liu2024visual, qin2024mp5, zhou2024minedreamer, qin2024worldsimbench, yin2023lamm, chen2024rh20t, ji2025robobrain, qin2025navigatediff, qi2025sofar, qin2024robofactory} greatly improve open-set \textit{reactive} failure detection for robotic.
Current LLM-based methods either convert visual inputs into text \cite{liu2023reflect, shinn2023reflexion, yao2022react}, potentially losing visual details, or rely on ground-truth feedback \cite{huang2022inner, silver2024generalized, skreta2023errors, raman2022planning}, which is impractical in real-world scenarios.
Recent studies employ VLMs as failure detectors, offering binary success indicators \cite{du2023vision, liu2024self, zheng2024evaluating, yu2023multireact} or textual explanations \cite{duan2024aha, dai2024racer} through visual question answering~(VQA), such as DoReMi~\cite{guo2023doremi}.
However, they often bear compromised execution speeds and coarse-grained detection accuracy.
% due to VLMs' high computational costs and limited 3D spatio-temporal perception capability.
%
Meanwhile, open-set \textit{proactive} failure detection is rarely explored, as previous methods~\cite{alvanpour2020robot, diehl2023causal} are confined to predefined failures.
%
% Meanwhile, open-set \textit{proactive} failure detection (\ie, preventing foreseeable failures) is rarely explored, as previous methods~\cite{alvanpour2020robot, diehl2023causal} are confined to predefined failures.
%
This detection mode requires foreseeing potential failure causes and monitoring them in real-time with high precision.
In this work, we formulate both \textit{reactive} and \textit{proactive} failure detection as constraint satisfaction problems and use VLM-generated code to evaluate them, meeting the expectations of both modes above.
%
% Subsequent works~\cite{guo2023doremi, yu2023multireact} detect intermediate failures during execution by frequently querying VLMs.
%
%
% , incurring high computational costs and increased hallucination risks; moreover, their reliance on images limits their ability to address failures requiring spatio-temporal reasoning.
%
%
% Furthermore, these VLM-based approaches rarely address \textit{proactive} failure detection (\ie, preventing foreseeable failures), and previous methods achieving proactive detection \cite{alvanpour2020robot, diehl2023causal} are confined to closed-set settings without leveraging LLMs/VLMs.

\vspace{+1mm}
\noindent\textbf{Visual Prompting.}
Visual prompts enhance the visual reasoning abilities of VLMs and are categorized into mask-based, point-based, and element-based methods. 
Mask-based approaches like SoM~\cite{yang2023set} apply numbered segmentation masks on images without considering instructions, whereas instruction-guided methods like IVM~\cite{zheng2024instruction} and API~\cite{yu2024attention} generate masks to block irrelevant regions. 
Point-based prompting encodes functionalities via points, including affordances~\cite{lee2024affordance, liu2024moka, yuan2024robopoint}, motion~\cite{nasiriany2024pivot, liu2024moka}, and constraints like ReKep~\cite{huang2024rekep}. 
However, ReKep~\cite{huang2024rekep} extracts semantic keypoints using DINOv2~\cite{oquab2023dinov2} and struggles to capture precise keypoints that accurately represent desired constraints. 
Element-based methods like CoPa~\cite{huang2024copa}, represent robotic control using basic elements (\eg, points, lines) via pre-trained models like SAM~\cite{kirillov2023segment} and GPT-4V~\cite{achiam2023gpt} simply. 
In contrast, this work explores constraint satisfaction monitoring and introduces constraint elements that more precisely represent relevant entities/parts.
These elements are tracked and evaluated in real-time to simplify monitoring.

\vspace{+1mm}
\noindent\textbf{Visual Programming.}
%
% Visual programming requires strong visual concept understanding and reasoning.
% %
% Prior works show strong generalization~(\eg, zero-shot) across various tasks, such as image editing~\cite{cho2024visual, gupta2023visual}, 3D visual grounding~\cite{yuan2024visual}, and robotics control~\cite{liang2023code, vemprala2024chatgpt, mu2024robocodex, venuto2024code}, by integrating LLMs with visual modules or VLMs.
% %
% However, these methods take raw RGB images as inputs with predefined primitive libraries.
% %
% % 
% In contrast, this work leverages constraint elements for visual programming and uses arithmetic operations in the code to encode their spatio-temporal combinational dynamics, presenting greater challenges.
%
% In contrast, we abstract constraints into various elements, annotate them onto the multi-view images, and encode their spatio-temporal relationships using arithmetic operations within the code, presenting greater challenges.
%
% This work introduces constraint-aware visual programming by leveraging VLMs with multi-view images annotated constraint elements.
% 
Visual programming requires strong visual concept understanding and reasoning.
Prior works show strong generalization~(\eg, zero-shot) across various tasks, such as image editing~\cite{cho2024visual, gupta2023visual}, 3D visual grounding~\cite{yuan2024visual}, and robotics control~\cite{liang2023code, vemprala2024chatgpt}, by integrating LLMs with visual modules but lose fine-grained details.
%but lose visual details
% primarily convert visual inputs into text for LLMs to process.
%
Recent studies~\cite{mu2024robocodex, venuto2024code} explore visual programming using VLMs, but these methods take raw RGB images as inputs with predefined primitive libraries.
In contrast, this work leverages constraint elements for visual programming and uses arithmetic operations in the code to encode their spatio-temporal combinational dynamics, presenting greater challenges.
%
% This work introduces constraint-aware visual programming by leveraging VLMs with multi-view images annotated constraint elements.
% 
\section{Method}
\label{sec: method}

\begin{figure*}[t]
\centering
\includegraphics[width=\linewidth]{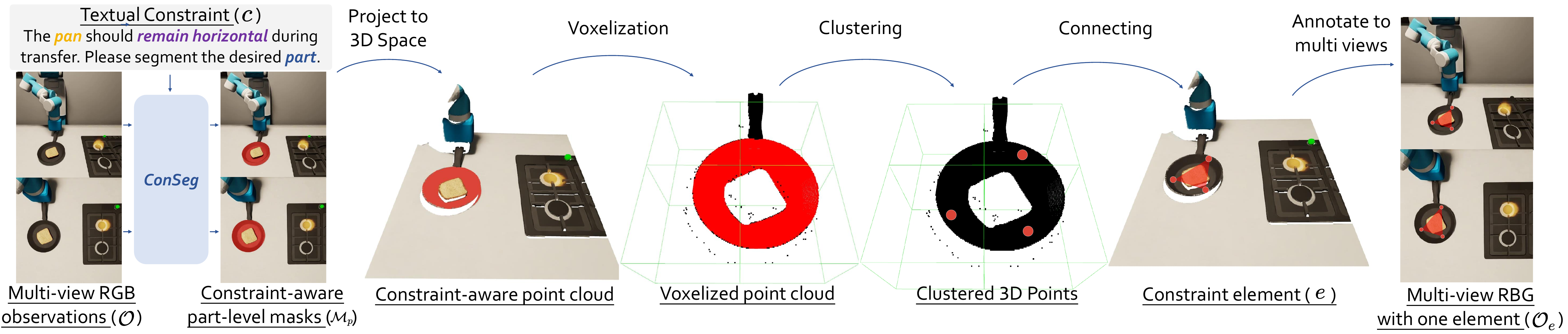}
\vspace{-6mm}
   \caption{Constraint Element Pipeline. Given a constraint, our model {\mname} generates instance-level and part-level masks across multiple views, which are projected into 3D space. 
   Through a series of heuristics, the desired elements are produced. Once all elements are obtained, they are annotated onto the original multi-view images. Here we display the annotation result of one element.}
\label{fig: element}
% \vspace{-5mm}
\vspace{-3mm}

\end{figure*}

We first give an overview of the proposed Code-as-Monitor (\fname)~(\cref{subsec: overview}). 
Then, we elaborate on the constraint element in Constraint Painter, especially constraint-aware segmentation~(\cref{subsec: constraint element proposal}).
Finally, we present Constraint Monitor for real-time detection~(\cref{subsec: real-time monitor module}). 
%
% , we give the necessary implementation details in \cref{{subsec: implementation details}}.

\subsection{Overview}
\label{subsec: overview}

The proposed {\fname} comprises three key modules: the Constraint Generator, Painter, and Monitor.
We focus on long-horizon manipulation task instructions $\mathcal{L}_{\mathrm{global}}$ (\eg, \textit{``Move the pan with the bread to the stove, and be careful not to let the bread fall out''}), using RGB-D observations $\mathcal{O}$ from two camera views~(front and top). 
As shown in \cref{fig: pipeline}, the RGB images~$\mathcal{O}$, along with instructions~$\mathcal{L}_{\mathrm{global}}$, previous subgoal~${l}_{\mathrm{pre}}$, and failure feedback from the Constraint Monitor~$f_{\mathrm{pre}}$~(\eg, subgoal success or failure reason), are fed into the Constraint Generator~$\mathcal{F}_{\mathrm{VLM}}$~(\ie, GPT-4o~\cite{achiam2023gpt}) to generate the next subgoal~${l}_{\mathrm{next}}$ and associated textual constraints~$\mathcal{C}$.
This process can be formulated as follows:

{
\vspace{-2mm}
\small
\begin{equation}
{l}_{\mathrm{next}}, \mathcal{C}_d, \mathcal{C}_u  = \mathcal{F}_{\mathrm{VLM}}(\mathcal{O}, \mathcal{L}_{\mathrm{global}}, {l}_{\mathrm{pre}}, {f}_{\mathrm{pre}} )
% \mathcal{C}_d = \{ {c}^{0}_d, {c}^{1}_d, \ldots, {c}^{i}_d \}\\ 
% \mathcal{C}_u = \{ {c}^{0}_u, {c}^{1}_u, \ldots, {c}^{i}_u \}
\tag{1}
\end{equation}
}

\noindent where $\mathcal{C}_d = \{ {c}^{0}_d, {c}^{1}_d, \ldots, {c}^{n}_d \}$ denotes the constraints that must be maintained during subgoal execution~(\eg, {pan handle must be grasped, bread must remain in the pan, pan should remain horizontal during transfer}), and $\mathcal{C}_u = \{ {c}^{0}_u, {c}^{1}_u, \ldots, {c}^{k}_u \}$ refers to the constraints that must be met upon subgoal completion~(\eg, {pan should be directly above the stove}).
We successfully unify \textit{reactive} and \textit{proactive} failure detection as these task-specific, situation-aware constraint satisfaction problems.

% These task-specific, situation-aware constraints determine the necessary spatio-temporal relationships for relevant entities or their parts, successfully unifying the open-set \textit{reactive} and \textit{proactive} failure detection.
% and simplify subsequent code generation.

%
% In Painter, for each textual constraint $c$ from $\mathcal{C}_d$ or $\mathcal{C}_u$, we generate corresponding constraint element proposals $e$~(detailed in \cref{subsec: constraint element proposal}) from RGB-D observations $\mathcal{O}$, which consist of 3D points in the world frame.
%
% These fine-grained, geometry-dependent elements represent the desired textual constraints~(\eg, the constant distance between green points on bread and pan determines if bread remains in pan, as shown in \cref{fig: pipeline}.)
In Painter, for each textual constraint $c$ from $\mathcal{C}_d$ or $\mathcal{C}_u$, we generate corresponding constraint elements $e$~(detailed in \cref{subsec: constraint element proposal}) from observations $\mathcal{O}$.
These elements, which are composed of 3D points, abstract the constraint-related entities or their parts to represent the desired textual constraint more easily~(\eg, the constant distance between green points on bread and pan determines if bread remains in pan, as shown in \cref{fig: pipeline}.)
These generated elements are then aggregated into the final set $\mathcal{E} = \{ {e}^{0}, {e}^{1}, \ldots, {e}^{n+k} \}$, and numerically annotated across all views to produce the final visual prompted images $\mathcal{O}_{\mathcal{E}}$.

In Monitor, we provide GPT-4o~\cite{achiam2023gpt} with the next subgoal~$l_{\mathrm{next}}$, textual constraints~$\mathcal{C}$, and annotated observations~$\mathcal{O}_{\mathcal{E}}$ for constraint-aware visual programming to generate the evaluation protocol, \ie, monitor code.
This code inputs the elements' 3D positions, calculates arithmetic operations within it, and returns a boolean to indicate potential or actual failure and a string to describe its reason. 
During subgoal execution, Monitor tracks the elements and evaluates the spatio-temporal combinational dynamics of these elements.
If the code returns \texttt{False}, the policy execution halts immediately, and the accompanying string is used as feedback \( f_{\mathrm{pre}} \) for re-planning. Otherwise, the subgoal is considered completed. In either case, the cycle is repeated.

\subsection{Constraint Element}
\label{subsec: constraint element proposal}
To simplify the monitoring of constraint satisfaction, we introduce constraint elements by abstracting constraint-related entities/parts into compact geometric elements~(\eg, points, lines) to get rid of the most irrelevant geometric and visual details, making them easier to track and generalize.

% To enhance the precision and efficiency of code generation and further computation, we introduce constraint elements by translating the constraints into combinable units, which encode the 3D spatio-temporal information of related constraints and are easy to track precisely.
%
% Generating monitor code directly from raw RGB images poses challenges for VLMs. To address this, we propose the constraint element, which encodes the 3D spatial information and textual constraints of entities within the images, serving as visual prompts to simplify the code generation.

\begin{figure}[t]
\centering
\includegraphics[width=\linewidth]{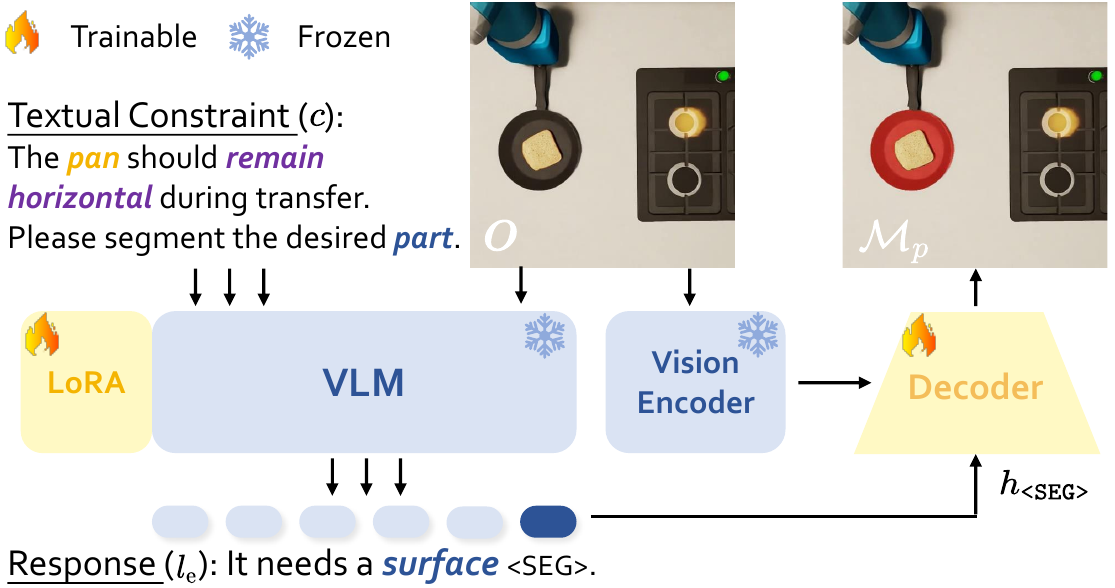}
\vspace{-6mm}
   \caption{{\mname} architecture. Here we display the part-level segmentation, which will output the desired element type and mask.
   % the annotation result of one element 
   % %
   % The VLM processes observation and textual constraint to output the element type and a \texttt{<SEG>} token. 
   % %
   % The decoder generates a mask based on the observation and \texttt{<SEG>}.
   % %
   % LoRA~\cite{hulora} is added to enable efficient fine-tuning.
   %
   }
\label{fig: network}
% \vspace{-5mm}
\vspace{-3mm}
\end{figure}

\vspace{+1mm}
\noindent \textbf{Pipeline.}
The constraint element generation pipeline is shown in {\cref{fig: element}}.
% For each textual constraint \( c \) from \( \mathcal{C}_d \) or \( \mathcal{C}_u \), we generate a unique element \( e \), as shown in {\cref{fig: element}}. 
%
Our trained multi-granularity constraint-aware model {\mname} performs two inference steps for each \( c \) and each RGB image \( o \) from the set of views \( O \). 
First, instance-level masks \( \mathcal{M}_{i} \) are generated to capture constraint-related entities. 
Then, fine-grained part-level masks \( \mathcal{M}_{p} \) and corresponding element type descriptions \( l_{\mathrm{e}} \) are produced, as shown in \cref{fig: network}.
Using corresponding depth data, we project \( \mathcal{M}_{p} \) from all views into 3D space, fusing them into a point cloud. 
%
% However, directly utilizing these entities or their parts in 3D space for code generation and computation is challenging due to the need for precise real-time tracking of their positions or poses and arithmetic computations within the code. 
%
However, directly tracking and evaluating the spatio-temporal combinational dynamics of these constraint-related entities/parts is challenging.
Therefore, we convert these entities into proposed constraint elements.
We first apply voxelization to the point cloud with voxel size determined by element type \( l_{\mathrm{e}} \)~(\eg, the surface needs at least $3$ points, we divide the occupied space into $2 \times 2$ voxels.).
Then we cluster one representative point within each voxel and filter them to a specified number, also determined by \( l_{\mathrm{e}} \) to extract the final desired 3D points. 
We connect points within each instance-level mask \( \mathcal{M}_i \) to form the constraint element \( e \) associated with constraint \( c \).
Notably, points modeled as end-effectors (\eg, the points of the fingertips and hand's center represent a dexterous hand) can be directly obtained from forward kinematics, bypassing the process above. 
Additionally, we perform parallel inference of \( \mname \) across all views to expedite the acquisition of the final set \( \mathcal{E} \) and their annotations onto the corresponding views \( \mathcal{O}_{\mathcal{E}} \).
Moreover, this minimalist approach, \ie, constraint elements, emphasizes the most relevant entities/parts, enabling generalization to unseen scenes and entities, which is critical for open-set failure detection.
%
% Overall, this minimalist approach emphasizes the most relevant entities/parts, offering potential diversity in the structured associations of constraint elements, which are crucial for generalizing unseen scenes and entities.
%  facilitating
% constraint satisfaction monitoring, 
% 
% bridging textual constraints with the monitor code, enabling precise real-time computation and facilitating \textit{proactive} failure detection.
%
For more details, please check Supp.~\ref{supsubsec: element pipeline details}.

% \vspace{+1mm}
\noindent \textbf{Constraint-aware Segmentation.}
%E
Since the textual constraint does not explicitly specify relevant entities/parts, we require a model capable of logical reasoning and certain open-set adaptation, enabling precise constraint-aware instance and part-level segmentation.
As depicted in {\cref{fig: network}}, we propose {\mname}, which builds upon LISA~\cite{lai2024lisa} comprising a VLM $\mathcal{F}_{\mathrm{VLM}}$ (\ie, LLaVA~\cite{liu2024visual}) with the vision encoder $\mathcal{F}_{\mathrm{enc}}$ and decoder $\mathcal{F}_{\mathrm{dec}}$ from SAM~\cite{kirillov2023segment}.
The textual constraint $c$ and the image $o$ are input into the VLM $\mathcal{F}_{\mathrm{VLM}}$ to generate an embedding for the \texttt{<SEG>} token.
The VLM outputs a textual description $l_{\mathrm{e}}$ of the desired element type additionally for part-level segmentation.
The decoder $\mathcal{F}_{\mathrm{dec}}$ then generates the segmentation mask $\mathcal{M}$ by leveraging visual features from the vision encoder $\mathcal{F}_{\mathrm{enc}}$ based on the current observation $o$, and features derived from the final-layer embedding $h_{\texttt{<SEG>}}$ of the \texttt{<SEG>} token, transformed via an MLP ($\mathcal{F}_{\mathrm{MLP}}$).
The process can be formulated as follows:

{
\vspace{-2mm}
\footnotesize
\begin{equation}
l_{\mathrm{e}}, h_{\texttt{<SEG>}} = \mathcal{F}_{\mathrm{VLM}}(c, o), \mathcal{M} = \mathcal{F}_{\mathrm{dec}}( \mathcal{F}_{\mathrm{MLP}}(h_{\texttt{<SEG>}}),  \mathcal{F}_{\mathrm{enc}}(o) )
\tag{2}
\end{equation}
}

% \noindent We adopt LISA's~\cite{lai2024lisa} loss function, including the next-token prediction loss for text output, and the combination of per-pixel BCE loss with DICE loss for mask output.
%
\noindent The entire pipeline, including additional LoRA~\cite{hulora} parameters, is optimized end-to-end while keeping the parameters of the VLM $\mathcal{F}_{\mathrm{VLM}}$ and vision encoder $\mathcal{F}_{\mathrm{enc}}$ frozen.
%
% For data collection and training details, please check Supp.~A.

\vspace{+1mm}
\noindent \textbf{Dataset and Collection Pipeline.} 
Our dataset utilizes images from BridgeData V2~\cite{walke2023bridgedata}. 
While BridgeData V2 provides trajectory-level instructions, we require frame-level annotations for per-frame subgoals and constraints.
To address this, we sample pick-and-place data, using external references~(\eg, gripper open/close states) to generate $10,181$ trajectories with $219,356$ images. 
We employ GPT-4o to decompose trajectory instructions into subgoals, constraints, and object-part associations to generate ground-truth annotations. 
Instance-level and part-level segmentations are performed using Grounded SAM~\cite{ren2024grounded} and Semantic SAM~\cite{li2023semantic}, respectively. 
These outputs are integrated and refined through manual inspection to produce the multi-granularity dataset, which is combined with LISA's training data to fine-tune our model. More details are in Supp.~\ref{supsec: implementation details}.

\subsection{Real-time Monitor Module}
\label{subsec: real-time monitor module}

After annotating constraint elements with unique colors and numerical labels across all views, these images serve as visual prompts to generate monitor code via GPT-4o~\cite{achiam2023gpt} with constraints $\mathcal{C}$ and the next subgoal $l_{\mathrm{next}}$. 
The generated code accepts the elements' 3D positions as input and performs arithmetic operations (\eg, NumPy-supported calculations) to evaluate spatio-temporal constraints at various stages (\ie, during execution or upon completion), as shown in \cref{fig: pipeline} code block. 
It returns a boolean flag indicating potential or actual failure if the computed result exceeds a predefined tolerance (\eg, accepting deviations of the pan's surface normal from the z-axis within $15^{\circ}$), along with a string explaining the cause. 
By tracking the elements using CoTracker~\cite{karaev23cotracker}, we achieve real-time detection via VLMs without frequent calls, reducing the computational cost.

Notably, including both current positions and historical trajectories of the elements as input allows us to monitor tasks needing high precision, such as the requirement of $2$cm movement or $180^{\circ}$ rotation, which previous \textit{reactive} detection methods could not address through direct VLM queries using raw RGB images. 
%
% Furthermore, the monitoring code can flexibly construct new implicit elements by recombining points from existing elements, enabling more precise computation of the desired constraints. 
%
%
Moreover, the potential diversity in the structured associations of constraint elements within the code~(\eg, calculating the angle between the normal vector of the pan's surface element and the z-axis determines if the pan is level, as shown in \cref{fig: pipeline} code block) and rich prior knowledge offered by VLMs enable generalization to certain unseen tasks requiring common skills.
%

% by expressing constraints as relationships between entities or parts, we achieve generalizability, allowing the abstraction of parts into elements to extend to common scenarios and objects, as shown in \cref{tab: hand 1}.

% Overall, this minimalist approach emphasizes the most relevant entities/parts, offering potential diversity in the structured associations of constraint elements, which are crucial for generalizing unseen scenes and entities.

% \subsection{Implementation Details}
% \label{subsec: implementation details}

% We use GPT-4o~\cite{achiam2023gpt} for all off-the-shelf VLM components and utilize CoTracker~\cite{karaev23cotracker, karaev24cotracker3} for real-time element tracking. \todo{training details and training dataset collection @CC.} 
% %
% For more data collection and training details about the {\mname} in Painter, please check Supp.~A.
\section{Experiments}
\label{sec: experiments}

Our experiments aim to address the following questions:
\textbf{(1)} Can {\fname} achieve open-set \textit{reactive} and \textit{proactive} failure detection across diverse robots, end-effectors, and objects, both in simulator~(\cref{subsec: main results in simulator}) and on real robots~(\cref{subsec: main results in real world})?
\textbf{(2)} Can {\mname} infer multi-granularity constraint-aware masks for unseen scenes and objects~(\cref{subsec: main results of segmentation})?
\textbf{(3)} Which design choices greatly enhance the performance~(\cref{subsec: ablation study})?

\subsection{Experimental Setup}
\label{subsec: experimental setup}
\textbf{Environment Settings.}
We evaluate {\fname} across three simulators~(\ie, CLIPort~\cite{shridhar2021cliport}, Omnigibson~\cite{li2022behavior}, and RLBench~\cite{james2019rlbench}) and a real-world setting.
In CLIPort, we employ a UR5 arm equipped with a suction cup for pick-and-place and a spatula for pushing, controlled by pre-trained CLIPort policy~\cite{shridhar2021cliport}. 
Omnigibson features a Fetch robot with a gripper, controlled by ReKep~\cite{huang2024rekep}. 
RLBench uses a Franka arm with a gripper, controlled by ARP~\cite{zhang2024autoregressive}.
% We collect a small dataset to fine-tune {\mname} for each simulator.
%
We use a UR5 arm with a Leap Hand~\cite{shaw2023leap} for real-world experiments, controlled by an open-loop policy named DexGraspNet 2.0~\cite{zhang2024dexgraspnet}.
More details are provided in Supp.~\ref{supsec: environment configuration}.

\vspace{+1mm}
\noindent \textbf{\mname~Configuration.}
Given the significant gap between simulation and real-world data, we fine-tune the \mname~model with $100$ trajectories collected from each simulation environment before deploying it. 
To share the same data collection and auto-label pipeline, this fine-tuning dataset is also limited to pick-and-place tasks. 
We manually annotate trajectory-level instructions, and image frames are captured at a frequency of $1$ Hz. 
Using only pick-and-place task data helps mitigate the data distribution gap while allowing us to validate the model's generalization to unseen tasks, including tool-use, articulated objects, and long-horizon complex tasks.
Notably, the \mname~model used in real-world experiments is not fine-tuned, ensuring a rigorous evaluation of \mname's generalization capability.

\vspace{+1mm}
\noindent \textbf{Task Settings.}
We introduce disturbances across all environments to induce failures, classified by the affected constraints (\eg, points, lines, surfaces).
We evaluate task success rate, execution time, and additional token usage in Omnigibson. 
Notably, we don't assess failure judgment success rate because our code operates continuously in real-time, rendering this metric inapplicable. 
Moreover, there is no standard for evaluating proactive failure detection accuracy. 
Detailed evaluation settings are provided in each section.

\vspace{+1mm}
\noindent \textbf{Baselines.}
We compare DoReMi~(DRM)~\cite{guo2023doremi} as the baseline for all settings, which uses VLMs to detect intermediate failures during execution via frequent VQA-based queries. 
In CLIPort, we include Inner Monologue~\cite{huang2022inner} baseline, which detects failures only upon each subgoal completion.

\subsection{Main Results in Simulator}
\label{subsec: main results in simulator}

\subsubsection{Results in CLIPort}
\label{subsubsec: cliport}

We evaluate two tasks in CLIPort:
\textbf{(1)} Stack in Order: The robot must stack blocks in a specified order, including two point-level disturbances: 
(a) with per-step probability \( p \), the suction cup may release a block, causing it to drop; 
(b) placement positions are perturbed by uniform noise in \([0, q]\) cm, potentially leading to tower collapse. 
Success is defined as correctly stacking the blocks within $70$s.
\textbf{(2)} Sweep Half the Blocks:
To address the pre-trained policy's tendency to sweep all blocks, the robot must determine when to halt, sweeping half of the blocks (±10\%) into a specified colored area within $30$s. 
Success is achieved by meeting these criteria.
\cref{tab: cliport} shows the mean results over $5$ different seeds, each with $12$ episodes.
For more details, please check Supp.~\ref{supsubsec: cliport evaluation details}.
The following paragraphs present our analyses.

\vspace{+1mm}
\noindent \textbf{Code better monitors 3D space relations.}
%
% As shown in the \cref{tab: cliport}, Under the most severe disturbances in ``Stack in Order'', {\fname} achieves a $75\%$ and $36\%$ higher success rate than those of MultiReact~\cite{yu2023multireact} and DoReMi~\cite{guo2023doremi}, respectively.
As shown in the \cref{tab: cliport}, Under the most severe disturbances~($p$=$0.3$, $q$=$3$) in ``Stack in Order'', {\fname} achieves a $17.5\%$ higher success rate than DoReMi.
Frequent VLM queries lead to increased incorrect failure judgments due to a limited 3D spatial understanding from single images, compared to code-based evaluation of block positions.
For example, after placing the green block on the red and preparing to pick up the blue, DoReMi mistakenly thinks that the green block is no longer atop the red, causing it to re-execute the previous subgoal.

\vspace{+1mm}
\noindent \textbf{Code with elements achieve \textit{reactive} and \textit{proactive} failure detetction.}
As shown in \cref{tab: cliport}, under severe disturbances in ``Stack in Order'', {\fname} reduces execution time by $38.7\%$ and $14.4\%$ compared to Inner Monologue~\cite{huang2022inner}, which only detects failures upon subgoal completion, and DoReMi~\cite{guo2023doremi}, which suffers from misjudgments due to repeated VLM queries, respectively.
In contrast, {\fname} unifies both \textit{reactive} and \textit{proactive} failure detection with high precision in real-time, especially in preventing foreseeable failures.
For example, if the green block is placed on the red block with heavy placement position disturbance, {\fname} anticipates that further stacking the blue may lead to collapse and then stabilizes the green block before proceeding.
%

% As shown in \cref{tab: cliport}, compared to Inner Monologue~\cite{huang2022inner}, which only detects failures at subgoal completion, and DoReMi~\cite{guo2023doremi}, which suffers from misjudgments due to repeated VLM queries during execution, {\fname} reducing execution time by $63\%$ and $29\%$, respectively, under the most severe disturbances~($p$=$0.3$, $q$=$3$) in ``Stack in Order''. 
% % 
% %
% The reason is that our model enables proactive failure detection, allowing us to prevent potential errors more efficiently.
% %
% For example, if the green block is placed on the red, {\fname} can detect that further stacking may lead to collapse due to placement position disturbance. Instead of placing a blue block on top, {\fname} would first stabilize the green block.

% As shown in the \cref{tab: cliport}, compared to Inner Monologue~\cite{huang2022inner}, which detects failures only at subgoal completion and ignores the failure during execution, and DoReMi~\cite{guo2023doremi}, which repeatedly queries the VLM during execution but suffers failure misjudgment, {\fname} detect failure with high precision in real-time, successfully reducing execution time by $63\%$ and $29\%$ under the most severe disturbances in ``Stack in Order'', respectively.
% 

\vspace{+1mm}
\noindent \textbf{Code with elements leads to more accurate counting.}
%
% As shown in the \cref{tab: cliport}, in ``Sweep Half the Blocks'', {\fname} achieves average success rates $6.4\times$ and $4.5\times$ higher than those of MultiReact~\cite{yu2023multireact} and DoReMi~\cite{guo2023doremi}, respectively. 
As shown in the \cref{tab: cliport}, in ``Sweep Half the Blocks'', {\fname} achieves average success rates $4.5\times$ higher than DoReMi~\cite{guo2023doremi}. 
Calculating the block points in the designated surface region provides more accurate results than directly using the VLM to count, enabling a more precise halting of the policy to complete the task.

\begin{table}[t]
\caption{Performance in CLIPort. We report the success rate and execution time, compared to CLIPort~(CP)~\cite{shridhar2021cliport}, with Inner Monologue~(IM)~\cite{huang2022inner} and DoReMi~(DRM)~\cite{guo2023doremi}.}
\vspace{-3mm}
\centering
\scriptsize
\setlength{\tabcolsep}{2.5pt}
\begin{tabular}{lc|c|ccc|ccc}
\bottomrule[1pt]
\multicolumn{2}{c|}{Tasks with}                  & \multicolumn{4}{c}{Success Rate(\%) $\uparrow$} & \multicolumn{3}{c}{Execution Time(s) $\downarrow$} \\
\multicolumn{2}{c|}{disturbance}                & CP & +IM  & +DRM & +\textbf{Ours}  & +IM & +DRM & +\textbf{Ours} \\
\hline
  Stack in         & $p$=0.0                   & 100.0 & 100.0  & 100.0 & 100.0    & 13.4 & 13.4 & 13.4   \\
  order with       & $p$=0.15                  & 56.67  & 81.67  & 83.33 & \textbf{95.00}        & 34.8 & 26.00 & \textbf{21.0} \\
  drop $p$         & $p$=0.3                   & 21.67 & 75.00  & 76.67 & \textbf{88.33}    & 42.8 & 34.20 & \textbf{25.4} \\
\hline
  Stack in         & $q$=1                     & 90.00  & 90.00  & 96.67 & \textbf{98.33}    & 24.8 & 24.6 & \textbf{24.2}   \\
  order with       & $q$=2                     & 41.67 & 71.67  & 75.00 & \textbf{83.33}        & 39.4 & 37.0 & \textbf{29.2} \\
  noise $q$        & $q$=3                     & 15.00 & 40.00  & 40.00 & \textbf{63.33}    & 58.2 & 54.2 & \textbf{36.8} \\
\hline
  \multicolumn{2}{c|}{Sweep Half the Blocks}   & 0.00 & 18.33  & 16.67 & \textbf{75.00}    & 22.0 & 16.6 & \textbf{16.4}   \\
\bottomrule[1pt]
\end{tabular}
\label{tab: cliport}
% \vspace{-5mm}
\vspace{-3mm}
\end{table}

\begin{table*}[t]
\caption{Performance in Omnigibson. We report the success rate~(SR), execution time, and token usage, compared to DoReMi~(DRM)~\cite{guo2023doremi}.}
\vspace{-3mm}
\scriptsize
\centering
\setlength{\tabcolsep}{3pt}
\begin{tabular}{l|cccc|cc|cccc|cc|cccc|cc}
\bottomrule[1pt]
\multirow{3}{*}{Method}        & \multicolumn{6}{c|}{Slot Pen (Point-level Disturbances)}                               
                               & \multicolumn{6}{c|}{Stow Book (Line-level Disturbances)}                           
                               & \multicolumn{6}{c}{Pour Tea (Surface-level Disturbances)}                          
                               \\

                               & \multicolumn{4}{c|}{SR with Disturbance(\%) $\uparrow$} & Time              &Token              & \multicolumn{4}{c}{SR with Disturbance(\%) $\uparrow$} & Time              &Token      
                               & \multicolumn{4}{c|}{SR with Disturbance(\%) $\uparrow$} & Time              &Token          
                               \\

                               & None & Dist.(a) & Dist.(b) & Dist.(c)     & (s)$\downarrow$   &(k)$\downarrow$  
                               & None & Dist.(a) & Dist.(b) & Dist.(c)     & (s)$\downarrow$   &(k)$\downarrow$   
                               & None & Dist.(a) & Dist.(b) & Dist.(c)     & (s)$\downarrow$   &(k)$\downarrow$ 
                               
                               \\
\hline
 ReKep~\cite{huang2024rekep}   & 30   & 20     & 10     &10          &  -                 &-           
                               & 40   & 30     &30      &20          &  -                 &-            
                               & 20   & 20   & 20      & 10          &  -                 &-           
             
                               \\
\hline
 ~~+DRM   & 40   & 10     & 20     &20          & 177.84                & 54.54          
                               & 50   & 40  & 20       &40          & 127.17                & 38.67   
                               & 0   & 0   & 0     & 0          &  -                    & -           

                               \\
 
 ~~\textbf{+Ours}              & \textbf{60}   & \textbf{50}     & \textbf{40}     &\textbf{40}          &  \textbf{101.85}                 & \textbf{25.82}           
                               & \textbf{70}   & \textbf{60}     &\textbf{70}      &\textbf{60}          &  \textbf{93.08}                  & \textbf{18.67}            
                               & \textbf{50}   & \textbf{40}     & \textbf{40}     & \textbf{30}          & \textbf{174.55}                & \textbf{44.19}            

                               \\

\bottomrule[1pt]
\end{tabular}
\label{tab: omnigibson}
% \vspace{-4mm}
\vspace{-2mm}
\end{table*}

\begin{figure*}[t]
\centering
\includegraphics[width=\linewidth]{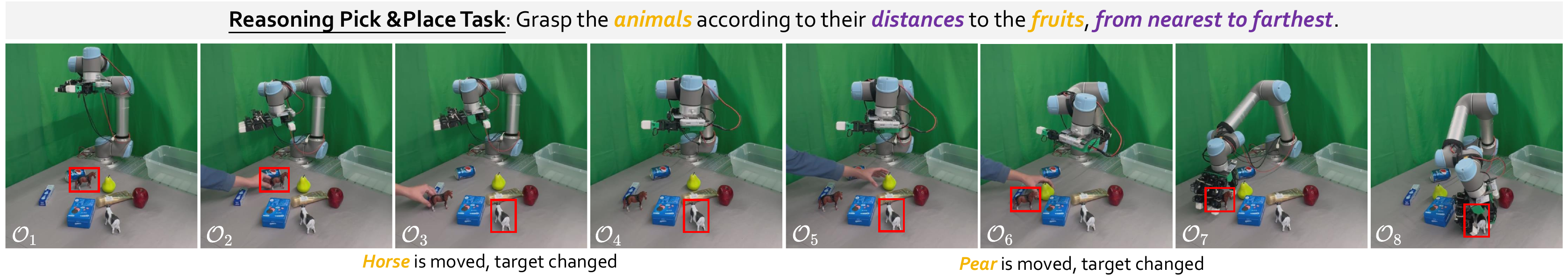}
\vspace{-7mm}
   \caption{Example of Real-world Evaluation.
   The red bounding box shows the current grasp target, which may shift due to environmental changes. {\fname} monitors and adapts to these changes in real-time, resulting in a closed-loop system with an open-loop policy.}
\label{fig: real_exp}
\vspace{-3mm}
\end{figure*}

\subsubsection{Results in Omnigibson}
\label{subsubsec: omnigibson}

We conduct experiments on three tasks in Omnigibson, each involving a distinct type of constraint-element disturbance:
\textbf{(1)} Slot Pen: Insert a pen into a holder, facing point-level disturbances wherein
(a) pen is moved during grasping;
(b) pen is dropped during transport;
(c) holder is moved during insertion.
\textbf{(2)} Stow Book: Place a book to a bookshelf vertically, with line-level disturbances where
(a) book is randomly rotated during grasping;
(b) end-effector joint is randomly actuated to alter the book pose;
(c) book is reoriented horizontally after placement.
\textbf{(3)} Pour Tea: Pours from a teapot into a teacup, encountering surface-level disturbances wherein
(a) teapot is tilted forward/backward during movement;
(b) end-effector joint is actuated to induce a lateral tilt of the teapot during movement;
(c) teapot is returned to a horizontal position during pouring.
\cref{tab: omnigibson} reports the results across three tasks, each including one no-disturbance trial and three specific-disturbance trials, with $10$ runs for each setting.
More details are provided in Supp.~\ref{supsubsec: omnigibson evaluation details}.
%
% Analyses are presented below.

\vspace{+1mm}
\noindent \textbf{Code with elements detects richer failures.}
%
% As shown in \cref{tab: omnigibson}, {\fname} surpasses the baseline across all tasks and disturbance types, with an average success rate improvement of $36\%$ over DoReMi~\cite{guo2023doremi}, which relies on VLM queries via image-based VQA. This demonstrates that {\fname} captures a broader range of failure cases involving diverse elements and their interactions. Notably, DoReMi’s success rate is lower than that of ReKep alone, due to its limited ability to handle spatio-temporal constraints when relying solely on a single image. For instance, in the "Pour Tea" task, DoReMi misinterprets deviations in the teapot's orientation, leading to incorrect re-planning and a 0% success rate.
%
In \cref{tab: omnigibson}, only {\fname} can detect failures caused by surface-level disturbances in ``Pour Tea'' compared to DoReMi~\cite{guo2023doremi}.
The reason is that changes in the teapot's pitch and roll angles are hard to detect by querying VLM via VQA using one current image.
% , which queries VLM solely on a single current image. 
%
The misjudgment leads to a $0\%$ success rate of DoReMi in this task.
% For example, in ``Pour Tea'', DoReMi misjudges deviations in the teapot's orientation or mistakenly assumes the robot is not holding the teapot due to occlusion, leading to incorrect re-planning and a $0\%$ success rate.
% {\fname} surpasses the baseline across all tasks and disturbance types, achieving an average success rate improvement of $36\%$ over DoReMi~\cite{guo2023doremi}.
%
% This shows that {\fname} can cover a richer set of failures involving diverse elements and their combination. 
%
Notably, DoReMi's success rate is sometimes lower than that of ReKep alone due to VLM's limited spatio-temporal understanding of single images. 

\begin{table}[t]
\caption{Performance of Single Pick \& Place. We report the success rate and execution time. DGN donates DexGraspNet 2.0~\cite{zhang2024dexgraspnet}.}
\vspace{-3mm}
\centering
\scriptsize
\setlength{\tabcolsep}{3pt}
\begin{tabular}{ll|c|cc|cc}
\bottomrule[1pt]
Tasks with         & Object       & \multicolumn{3}{c}{Success Rate(\%) $\uparrow$} & \multicolumn{2}{c}{Execution Time(s) $\downarrow$} \\
disturbance           & types               & DGN  & +DRM & +\textbf{Ours}   & +DRM & +\textbf{Ours} \\
\hline
 Pick \& Place  & Deformable       & 0.00  & 83.33 & \textbf{96.67}        & 61.8 & \textbf{46.3}   \\
 with the       & Transparent      & 0.00  & 66.67 & \textbf{93.33}        & 72.6 & \textbf{48.1} \\
 objects being  & Small Rigid      & 0.00  & 80.0 & \textbf{96.67}        & 65.7 & \textbf{45.4} \\
 moved          & Large Geometric  & 0.00  & 86.67 & \textbf{96.67}        & 68.9 & \textbf{45.3} \\
  
\hline
 Pick \& Place  & Deformable       & 0.00  & 76.67 & \textbf{93.33}        & 68.7 & \textbf{62.5}   \\
 with the       & Transparent      & 0.00  & 60.00 & \textbf{90.00}        & 77.7 & \textbf{62.7} \\
 objects being  & Small Rigid      & 0.00  & 63.33 & \textbf{93.33}        & 69.8 & \textbf{60.5} \\
 removed     & Large Geometric  & 0.00  & 76.67 & \textbf{96.67}        & 72.3 & \textbf{60.3} \\
\bottomrule[1pt]
\end{tabular}
\label{tab: hand 1}
% \vspace{-2mm}
\end{table}

\begin{table}[t]
\caption{Performance of Reasoning Pick \& Place in cluttered scene. We report the success rate. 
The robot is controlled by an open-loop policy named DexGraspNet 2.0~(DGN)~\cite{zhang2024dexgraspnet}.
w/ CE with $\checkmark$ indicates using constraint elements; otherwise, constraint-aware entities or parts are used for tracking and code computation.}
\vspace{-3mm}
\centering
\scriptsize
\setlength{\tabcolsep}{3pt}
\begin{tabular}{l|c|c|cc}
\bottomrule[1pt]
\multirow{2}{*}{Tasks}      &  \multirow{2}{*}{w/ CE}              & \multicolumn{3}{c}{Success Rate(\%) $\uparrow$} \\
                            &               & DGN & +DRM & +\textbf{Ours} \\
\hline
\multirow{2}{*}{Clear all objects on table except for animals} & \checkmark  &  {0.00}  & {10.00} & \textbf{60.00} \\      
                                                                   & \ding{55}  &  {0.00}  & {10.00} & \textbf{20.00} \\     
\hline
Grasp the animals according to their  & \multirow{2}{*}{\checkmark}  &  \multirow{2}{*}{0.00} & \multirow{2}{*}{0.00} & \multirow{2}{*}{\textbf{90.00}} \\      
distances to fruits, from nearest to farthest  &  &                     &                       &                  \\

\bottomrule[1pt]
\end{tabular}
\label{tab: hand 2}
% \vspace{-6mm}
\vspace{-3mm}
\end{table}

\vspace{+1mm}
\noindent \textbf{Code with elements detects failure with lower computational cost.}
As shown in \cref{tab: omnigibson}, we achieve a $34.8\%$ reduction in execution time and a $52.2\%$ decrease in token count compared to DoReMi~\cite{guo2023doremi}. 
% 
% This improvement stems from combining \textit{reactive} and \textit{proactive} failure detection, which 可以实时精确的发现错误并及时re-plan防止出现更严重的failure，同时 generating code only once per subgoal.
%
This improvement stems from \textit{proactive} failure detection, which prevents more severe failures ahead in real-time for timely re-planning while generating code only once per subgoal.
% This improvement stems from combining \textit{reactive} and \textit{proactive} failure detection, enabling precise identification of occurred failures and real-time prevention of potential ones to re-plan while generating code only once per subgoal.
% 
%
For example, in ``Pour Tea'', {\fname} detects failure by monitoring the teapot's tilt angle in real-time, avoiding frequent VLM checks.

% As shown in \cref{tab: omnigibson}, we achieve a $50\%$ reduction in execution time and a threefold decrease in token number compared to DoReMi~\cite{guo2023doremi}.
% %
% The former benefits from the combination of \textit{reactive} and \textit{proactive} failure detection, while the latter is due to generating code only once at the start of each subgoal.
% %
% For instance, in ``Pour Tea'', when the teapot deviates from the \( x\text{-}z \) plane, {\fname} successfully predicts potential failure by calculating the teapot's real-time tilt angle and adjusts the robot's trajectory accordingly, without frequent querying VLM.
%

\vspace{-2mm}
\subsubsection{Results in RLBench}

We further evaluate {\fname} on RLbench and demonstrate its superior generalization across diverse manipulation tasks, including articulated objects, rotational manipulation, and tool use.
Experimental details can be found in Supp.~\ref{supsubsec: rlbench evaluation details}.

% \label{subsubsec: rlbench}
%  We conducted 6 different experiments of 3 types on RLbench to demonstrate the robustness and effectiveness of our model.  The tasks were designed to test the robot's ability to handle complex manipulations involving articulated objects and tool use, which require precise adherence to constraints for successful execution.:
%  %
% \textbf{(1)} Articulated Object: 
% %
% (a) open drawer: The robot must grasp and pull open a drawer,; 
% %
% (b) put in drawer: The robot must open the drawer and then put an item into it. 
% %
% \textbf{(2)} Tool Use:
% %
% (a) Screw Bulb: The robot needs to rotate a bulb into a socket; 
% %
% (b) Turn Tap: Turning a faucet tap. 
% %
% \textbf{(3)} Tool Use:
% %
% (a) Drag Stick; 
% %
% (b) Sweep to Dustpan. 
% %
% \cref{tab: rlbench} reports the results across six tasks, we compared {\fname} with sota models RVT2 and ARP with $1000$ episode runs per task. Additionally, we did not use the Faster Point-Renderer in our environment which pre-trained models are trained with.
% %
% More details are provided in Supp.~C.2.
% %
% Analyses are presented below.

% \vspace{+1mm}
% \noindent \textbf{Code can generalize better to monitor diverse tasks.}
% As shown in the \ref{tab: rlbench}, {\fname} significantly enhances the robot's ability to perform complex manipulation tasks with 7.37\% higher success rates compared to ARP. By integrating constraint elements and code-based monitoring, {\fname} provides precise real-time computation of spatial and temporal constraints for both reactive and proactive failure detection.

\begin{table}[t]
\caption{Performance of reasoning and constraint-aware segmentation. FMC denotes the foundation model combination baseline.}
\vspace{-3mm}
\centering
\scriptsize
\begin{tabular}{l|cc|cc|cc}
\bottomrule[1pt]
\multirow{3}{*}{Method}  & \multicolumn{2}{c|}{ReasonSeg} & \multicolumn{4}{c}{ConstraintSeg} \\
% \cline{2-7}
 & \multicolumn{2}{c|}{Instance-level} & \multicolumn{2}{c|}{Instance-level} & \multicolumn{2}{c}{Part-level}\\
% \cline{2-7}
& gIoU & cIoU  & gIoU & cIoU  & gIoU & cIoU  \\
\hline
  OVSeg~\cite{liang2023open}            & 28.5  & 18.6  & 32.9 & 31.4  & 20.2 & 21.5 \\
  GRES~\cite{liu2023gres}               & 22.4  & 19.9  & 28.6 & 26.4  & 22.7 & 22.6 \\
  X-Decoder~\cite{zou2023generalized}   & 22.6  & 17.9  & 27.8 & 28.0  & 23.5 & 25.1 \\
  SEEM~\cite{zou2024segment}            & 25.5  & 21.2  & 29.4 & 28.0  & 23.1 & 24.8 \\
  PixelLM~\cite{ren2024pixellm}         & 56.0  & 61.4  & 44.4 & 43.2  & 24.1 & 22.6  \\
  LISA-13B~\cite{lai2024lisa}           & \textbf{56.6}  & \textbf{65.1}  & 42.1 & 38.9  & 23.4 & 24.3  \\
  FMC                                   & 51.2  & 53.3  & 49.3 & 49.6  & 40.8 & 39.3 \\
\hline
  \textbf{{\mname}-13B}   & 55.7  & 63.9  & \textbf{62.1} & \textbf{68.7}  & \textbf{60.2} & \textbf{65.3} \\
\bottomrule[1pt]
\end{tabular}
\label{tab: conseg}
% \vspace{-6mm}
\vspace{-3mm}
\end{table}

\begin{figure*}[t]
\centering
\includegraphics[width=\linewidth]{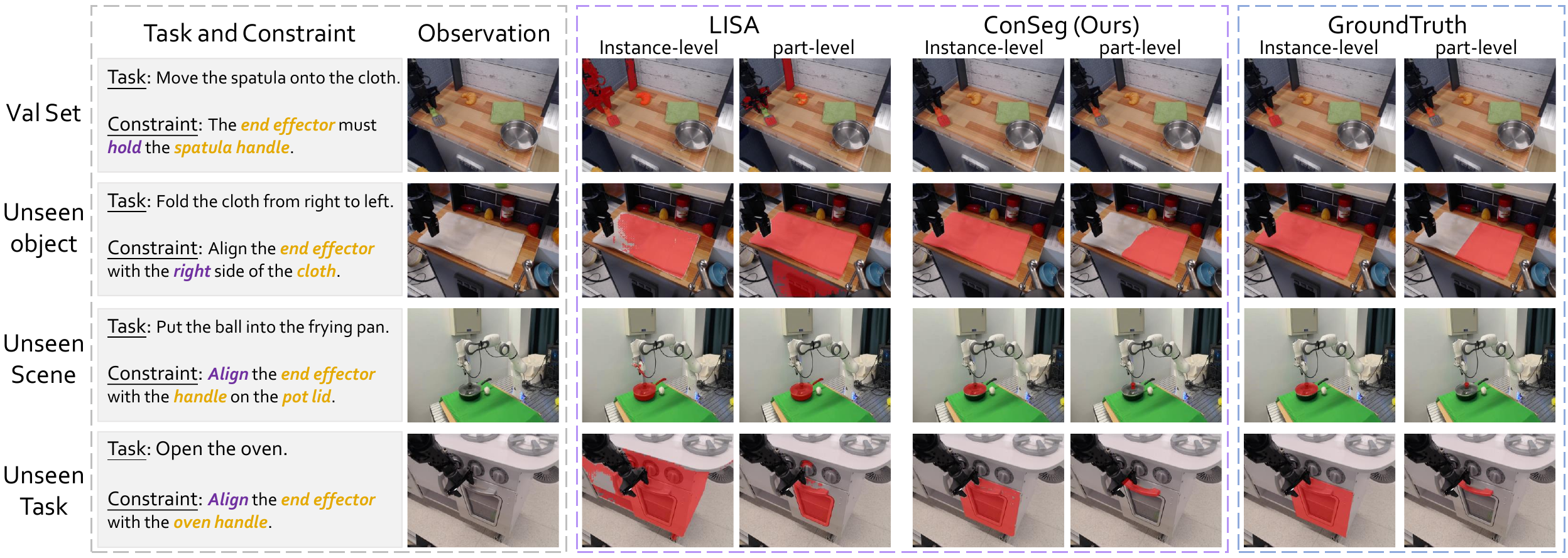}
\vspace{-6mm}
   \caption{Visual comparison between our {\mname} and LISA~\cite{lai2024lisa} at instance and part level. The red masks are the segmentation results.}
\label{fig: seg}
% \vspace{-5mm}
\vspace{-3mm}
\end{figure*}

\subsection{Main Results in Real World}
\label{subsec: main results in real world}

We conduct real-world evaluations on two tasks:
\textbf{(1)} Simple Pick \& Place: The robot has $70$s to pick up objects and place them at specified locations, facing two disturbances: (a) moving the object during grasping, and (b) removing the object from hand during movement. 
We test four object types (\eg, deformable, transparent), selecting $3$ examples per type and conducting $10$ trials for each (see \cref{tab: hand 1}). 
\textbf{(2)} Reasoning Pick \& Place: The robot executes long-horizon tasks, involving ambiguous terms (\eg, ``fruit'', ``animal''), under the same disturbances.
We evaluate $2$ long-horizon tasks in cluttered scenes, performing $10$ trials each (see \cref{tab: hand 2}).
%
% Notably, the robot is controlled by an open-loop policy, \ie, DexGraspNet 2.0.
%
For more details, please check Supp.~\ref{supsubsec: real-world evaluation details}.
The following paragraphs present our analyses.

\vspace{+1mm}
\noindent \textbf{Elements generally abstract constraints and relevant entities.}
%
% As shown in \cref{tab: hand 1}, using a different end-effector (\ie, Leap Hand) in Simple Pick \& Place, {\fname} also achieves success rates surpassing MultiReact~\cite{yu2023multireact} and DoReMi~\cite{guo2023doremi} by $77.0\%$ and $22.3\%$, respectively.
As shown in \cref{tab: hand 1}, using a different end-effector (\ie, Leap Hand~\cite{shaw2023leap}) in Simple Pick \& Place, {\fname} also achieves success rates surpassing DoReMi~\cite{guo2023doremi} by $20.4\%$ when handling different kinds of objects~(\eg, deformable).
We find that abstracting constraint-related entities/parts removes the irrelevant visual details, enabling generalization to different kinds of entities in unseen scenes~(as discussed in \cref{subsec: constraint element proposal}), leading to easier tracking and code evaluation.

% We find that abstracting constraints as relationships between entities or parts provides generalizability, and further converting them into elements can better model end-effectors and objects dependent on the element type, geometric, and \etc, as discussed in \cref{subsec: constraint element proposal}, leading to easier tracking and code computation.
%
% For example, the dexterous hand can be modeled as five points at the fingertips and the hand center, with positions derived from forward kinematics easily.

% \vspace{+1mm}
\noindent \textbf{\textit{Reactive} and \textit{Proactive} failure detection combined with an open-loop policy forms a close-loop system.}
\cref{tab: hand 2} shows that only {\fname} successfully handles long-horizon tasks in cluttered scenes, while all baselines fail. 
These tasks are challenging as the robot is controlled by an open-loop policy that cannot handle environment dynamics and human disturbances in a closed-loop manner.
By incorporating both \textit{reactive} and \textit{proactive} failure detection with the open-loop policy, as shown in \cref{fig: real_exp}, the robot can dynamically adjust its target object in real-time.
For example, when a human moves the horse or pear during the task, the robot adapts by grasping the animal closest to the fruit, effectively forming a closed-loop system.

% \cref{tab: hand 2} shows that only {\fname} successfully performs long-horizon tasks in cluttered scenes, while all baselines basically fail.
% %
% These tasks are quite difficult as the robot is controlled by an open-loop policy, \ie, DexGraspNet 2.0~\cite{zhang2024dexgraspnet}, which cannot deal with environment dynamics and human disturbance in a close-loop manner.
% %
% Thanks to integrating the \textit{reactive} and \textit{proactive} failure detections with the open-loop policy, as depicted in \cref{fig: real_exp}, the robot can dynamically adjust its target object in real-time precisely.
% %
% For example, when the human moves the horse or pear during execution, the robot adapts to environmental changes and grasps the nearest fruit among animals, thereby achieving a closed-loop system.

\begin{table}[t]
\caption{
Ablation studies in Omnigibson's ``Stow Book'', assessing the impact of Multi-View (MV), Constraint-aware Segmentation (CS) for elements, and Connect Points (CP) for element formation. 
% $\checkmark$ indicates inclusion; a \ding{55} indicates omission.
}
\vspace{-3mm}
\scriptsize
\centering
\begin{tabular}{ccc|ccccc}
\bottomrule[1pt]                           
\multirow{2}{*}{MV} & \multirow{2}{*}{CS} & \multirow{2}{*}{CP} & \multicolumn{5}{c}{Success Rate with Disturbance(\%) $\uparrow$} \\
                            &                             &                                & None & Dist.(a) & Dist.(b) & Dist.(c) &Avg  \\
\hline

\ding{55} & \checkmark & \checkmark    & 40.0   & 40.0     & 30.0     &50.0    &40.0     \\

\checkmark & \ding{55} & \checkmark    & 50.0   & 40.0     & 40.0     &40.0    &42.5 \\
 
\checkmark & \checkmark & \ding{55}    & 60.0   & 50.0     & 60.0     &50.0    &55.0   \\

\checkmark & \checkmark & \checkmark   & \textbf{70.0}   & \textbf{60.0}     & \textbf{70.0}     &\textbf{60.0}  &\textbf{65.0}      \\
\bottomrule[1pt]
\end{tabular}
\label{tab: ablation 1}
% \vspace{-5mm}
\vspace{-3mm}
\end{table}

\subsection{Main Results of Segmentation}
\label{subsec: main results of segmentation}

\cref{tab: conseg} presents segmentation results comparing our {\mname}, with SOTA models and a Foundation Model Combination (FMC) baseline. 
FMC integrates GPT-4o for reasoning over tasks and constraints to identify relevant instances and parts, along with Grounded SAM~\cite{ren2024grounded} and Semantic SAM~\cite{li2023semantic} for instance and part-level segmentation, respectively, similar to our data collection pipeline. 
We evaluate performance on the ReasonSeg~\cite{lai2024lisa} benchmark and our proposed Constraint-Aware Segmentation (ConstraintSeg) benchmark. 
Our benchmark evaluates performance using gIoU and cIoU, following ReasonSeg's evaluation setting.

\vspace{+1mm}
\noindent \textbf{{\mname} performs both reasoning and multi-granularity constraint-aware segmentation.}
As shown in \cref{tab: conseg}, {\mname} performs comparably to LISA and PixelLM on the ReasonSeg but significantly surpasses them on ConstraintSeg, achieving nearly a $40\%$ improvement at the part level. Visual comparisons in \cref{fig: seg} further demonstrate {\mname}'s strong generalization to unseen objects, scenes, and tasks.

\subsection{Ablation Study}
\label{subsec: ablation study}

% We conduct ablation studies and present analyses below.

We conduct ablation studies following the corresponding environment settings and present analyses below.
% %
% The analyses are presented below.

\vspace{+1mm}
\noindent \textbf{Multi views are critical for visual programming.}
% As shown in \cref{tab: ablation 1}, using only a front-view image with constraint elements reduces the average success rate from $65\%$ to $40\%$, because accurately determining lines and surfaces requires multiple viewpoints for visual programming; a single view can lead to occlusion or dimensional reduction, causing surfaces to be perceived as lines or lines as points.
% As shown in \cref{tab: ablation 1}, using only a front-view image with constraint elements reduces the average success rate from $65\%$ to $40\%$. 
%
As shown in \cref{tab: ablation 1}, using only front-view images with constraint elements reduces the average success rate from $65\%$ to $40\%$. This is primarily due to (1) occlusion leading to suboptimal element generation and (2) errors in visual programming, where dimensional reduction causes surfaces to be misinterpreted as lines or lines as points.

%这是因为(1) occlusion 会生成较差的 element，(2) visual programming 会识别element错误，出现dimensional reduction，causing surfaces to be perceived as lines or lines as points.

\begin{table}[t]
\caption{Ablation study on training data. SemanticSeg includes ADE20K~\cite{zhou2017scene}, COCO-Stuff~\cite{caesar2018coco}, PACO-LVIS~\cite{ramanathan2023paco} and PASCAL-Part~\cite{chen2014detect}. ReferSeg includes refCLEF, refCOCO, refCOCO+~\cite{kazemzadeh2014referitgame} and refCOCOg~\cite{mao2016generation}. VQA indicates LLaVAInstruct-150k~\cite{liu2024visual}. }
\vspace{-3mm}
\scriptsize
\centering
\setlength{\tabcolsep}{1.5pt}
\begin{tabular}{cccccc|cccc}
\bottomrule[1pt]
\multirow{2}{*}{SemanticSeg} & \multirow{2}{*}{ReferSeg} & \multirow{2}{*}{VQA} & \multirow{2}{*}{ReasonSeg}  & \multicolumn{2}{c|}{ContraintSeg} & \multicolumn{2}{c}{Ins-level}  & \multicolumn{2}{c}{Part-level}\\                           
& & & & Ins & Part & gIoU & cIoU & gIoU & cIoU   \\
\hline
\checkmark & \checkmark & \checkmark & \checkmark & \ding{55} & \ding{55} & 42.1 & 38.9 & 23.4 & 24.3\\
\checkmark & \checkmark & \checkmark & \checkmark & \checkmark & \ding{55} & 60.7 & 65.9 & 40.4 & 45.6\\
\checkmark & \checkmark & \checkmark & \checkmark & \ding{55} & \checkmark & 51.5 & 50.6 & 56.5 & 61.7\\
\checkmark & \checkmark & \checkmark & \checkmark & \checkmark & \checkmark & \textbf{62.1} & \textbf{68.7} & \textbf{60.2} & \textbf{65.3} \\

\bottomrule[1pt]
\end{tabular}
\label{tab: ablation data}
% \vspace{-5mm}
\vspace{-3mm}
\end{table}

\vspace{+1mm}
\noindent \textbf{Constraint-aware segmentation enhances element.}
As shown in \cref{tab: ablation 1}, using DINOv2~\cite{oquab2023dinov2} to generate semantic points to form elements, rather than through constraint-aware segmentation, reduces the success rate from $65\%$ to $42.5\%$, because these constructed elements fail to represent the desired constraints accurately.
For example, capturing a book's vertical orientation requires two precise points on its edge, which DINOv2 can not provide.
% As shown in \cref{xxx}, using DINOv2~\cite{oquab2023dinov2} to generate semantic points compared to constraint-aware segmentation for forming the elements reduces the success rate from XX to XX due to the constructed elements' inability to accurately represent the desired constraints (\eg, using two correct points on the book's edge to form a line describing its vertical orientation).

\vspace{+1mm}
\noindent \textbf{Forming elements improves code generation.}
As shown in \cref{tab: ablation 1}, using unconnected 3D points as final elements reduces the success rate from $65\%$ to $55\%$. 
Pre-formed elements contain more prior constraint information, serving as visual cues better encoded in code, whereas standalone points require recombination during visual programming.

\vspace{+1mm}
\noindent \textbf{Elements simplify constraints computation.}
% tracking and computation in a real-world setting
%
As shown in \cref{tab: hand 2}, the success rate decreases from $60\%$ to $20\%$ when entities or parts are not transformed into proposed elements in real-world evaluation. 
We find two key issues: 
(1) inaccurate 3D position and slow pose tracking of entities or parts; and 
(2) difficulties in performing arithmetic operations on their 3D positions or poses in code, hindering the encoding of their spatio-temporal constraints.

\vspace{+1mm}
\noindent \textbf{Both instance-level and part-level data improve performance.}
We conduct ablation studies to assess the impact of our proposed dataset on the ConstraintSeg benchmark. The results are shown in \cref{tab: ablation data}. 
The results indicate that both the instance-level and part-level subsets contribute to performance improvements and enhance each other's effects.

\section{Conclusion}
\label{sec: conclusion}
In this paper, we present a novel paradigm termed Code-as-Monitor, leveraging the VLMs for both open-set reactive and proactive failure detection.
In detail, We formulate both detection modes as spatio-temporal constraint satisfaction problems and use VLM-generated code to evaluate them for real-time monitoring.
We further propose constraint elements, which abstract constraints-related entities or their parts into compact geometric elements, to improve the precision and efficiency of monitoring. 
Extensive experiments demonstrate the superiority of the proposed approach and highlight its potential to advance closed-loop robot systems.
%
% we encode the spatio-temporal constraints into constraint elements as visual prompts and generate monitor code to check in real-time more accurately.
%
% A multi-granularity constraint-aware segmentation model is further developed to obtain the constraint-related instance and part to better form the elements.
%

% \vspace{+1mm}
% \noindent\textbf{Limitation and Future Work.}
% %
% The proposed approach still has the following limitations. 
% % Despite its effectiveness, the proposed approach has limitations. 
% %
% First, it relies on stable and accurate 3D point tracking; however, our approach, \ie, using a 2D point tracker, makes it susceptible to occlusion-induced inaccuracies. Employing a 3D tracker could mitigate this issue. 
% %
% Second, the method focuses on failures detectable through displacement, making it less effective for detecting failures related to force direction in real-time, which is a potential area for further research.

\section*{Acknowledgement}
This work was supported by the National Natural Science Foundation of China (62132001), the National Science and Technology Major Project (2022ZD0116314), and the Fundamental Research Funds for the Central Universities.

\noindent We sincerely thank Xingqiang Yu for his valuable discussions and insightful feedback about the Leap Hand and UR5 in real-world experiments.
We also sincerely appreciate Jingyi Yang's excellent figure design (\eg, teaser figure, pipeline overview) and demo editing work.

% \newpage

{
    \small
    \bibliographystyle{ieeenat_fullname}
    \bibliography{main}
}

% WARNING: do not forget to delete the supplementary pages from your submission 
% \input{sec/X_suppl}
\clearpage
\appendix
\setcounter{page}{1}
\maketitlesupplementary

\noindent The supplementary document is organized as follows:
\newline
\begin{itemize}

    \item Sec.~\ref{supsec: more discussion}: More Discussions like Technical Details, Limitations, and Future Work.
    \newline
    \item Sec.~\ref{supsec: implementation details}: Implementation Details of Painter, including data collection, training, and element pipeline details.
    \newline
    \item Sec.~\ref{supsec: environment configuration}: Environment Configuration, including environmental setups, control policies, and baseline details.
    \newline
    \item Sec.~\ref{supsec: evaluation details}: Evaluation Details, including detailed task definition, disturbances, and experimental results.
    \newline
    \item Sec.~\ref{supsec: more ablation}: More Ablation Studies.
    \newline
    
    \item Sec.~\ref{supsec: more demonstrations and prompts}: More Demonstrations of {\fname}.
\end{itemize}

\section{More discussions}
\label{supsec: more discussion}

\subsection{Methodology discussions}

\textbf{What contributes to generalization.}
VLM's intrinsic knowledge and reasoning enable task-level generalization, while geometric abstraction ensures generalization across scenes and entities (\eg, objects, robots). 
The constraint-based formulation seamlessly integrates both aspects, achieving overall framework generalization.

\vspace{+1mm}
\noindent\textbf{Proactive detection for long-term dynamics.}
Our framework can potentially handle long-term dynamics by leveraging the Constraint Generator to decompose long-horizon tasks into subgoals and enables proactive detection within each subgoal to \textit{constrain the future state space} rather than directly predict future states.
Moreover, failed subgoals are re-planned with updated proactive constraints, ensuring adaptability to long-term dynamic changes.

\vspace{+1mm}
\noindent\textbf{Key methodology compared to Rekep~\cite{huang2024rekep}.}
Our key method differs from ReKep in two key aspects:
\textbf{(1) Element extraction:} ReKep simply relies on DINOv2 for keypoint detection via 2D feature clustering, which offers a limited representation of constraints and geometric relationships crucial for failure detection.
In contrast, our method achieves higher-order geometric abstraction by leveraging {\mname} to extract constraint-aware objects/parts in 2D space, then combining these with point clouds to integrate 3D geometric constraints.
\textbf{(2) Code Generation:} Unlike ReKep, which lacks fine-grained guidance, our method leverages these \textit{geometry-rich elements} as visual prompts on \textit{multi-view} images to generate code more accurately \textit{at each subgoal's beginning}, enabling seamless integration with various policies to form closed-loop systems.

\vspace{+1mm}
\noindent\textbf{Part segmentation for ambiguous or incomplete instructions.}
In our framework, we mainly leverage {\mname}'s abilities to handle cases where instructions with ambiguous or missing explicit part information.
\textbf{(1)} We do not impose whether the textual instructions contain part information in our dataset collection pipeline. 
This allows the {\mname} to possess a certain ability to handle ambiguous instructions.
\textbf{(2)} {\mname} architecture adopts a VLM-based design, leveraging VLM’s world knowledge to infer constraint-related parts for unseen objects.
Based on our experiments (see Supp.~\ref{supp: constraint-aware segmentation demo}), {\mname} does demonstrate some generalization ability, but its performance on unseen objects is weaker than on objects within the training distribution.

\subsection{Technical detail discussions}

\textbf{Executability and reliability of code.}
For executability, we validate the code using the path coverage of White-box Testing at the subgoal's beginning, where the current state is used as input to verify the accuracy of each logic branch (\ie, every if-else block). If any error occurs during testing, the code is regenerated.
For reliability, we prompt VLM to generate code that closely adheres to the given specifications and requirements. 
As frequent VLM calls increase hallucinations, we only invoke VLM once at each subgoal's beginning, making it less affected by hallucination and more reliable.

\vspace{+1mm}
\noindent\textbf{Failure threshold Selection.}
We use two ways: 
\textbf{(1)} manually initializing thresholds for common tasks in an external knowledge base; 
\textbf{(2)} generating thresholds for unseen tasks using VLM's internal knowledge. 
We find that the system remains robust to threshold selection in common settings, but the impact of threshold variations under diverse conditions remains unexplored and is left for future work.

\vspace{+1mm}
\noindent\textbf{Computational cost.}
The segmentation model is only invoked at the subgoal's beginning to generate elements and \textit{excluded during execution}, ensuring real-time code-based monitoring by tracking only the elements without high computational cost. 
We also perform parallel inference of segmentation models to accelerate at the subgoal's beginning.

\subsection{Limitations and Future Work}
Despite promising results, our approach has several limitations.
First, using Visual Language Models (VLMs)—such as off-the-shelf GPT for code generation and constraint-aware segmentation models—inevitably leads to hallucination issues. Even with minimized VLM usage, inaccuracies in code generation and biases in segmentation may persist. Integrating more relevant knowledge via methods like Retrieval-Augmented Generation (RAG)~\cite{lewis2020retrieval} or reducing multimodal large language model (MLLM) hallucinations~\cite{MinerviniAHD2024} could alleviate these problems.
Second, while we unify reactive and proactive failure detection as spatio-temporal constraint satisfaction problems and propose constraint elements for simplified real-time, high-precision monitoring, the constraint element representation has limitations:
(1) It primarily focuses on failures detectable through explicit displacement and rotation, rendering it less effective for force-direction-related failures without noticeable displacement—for example, a robotic gripper failing to open a drawer due to incorrect force application.
(2) The simplified representation abstracts objects and minimizes irrelevant visual details for efficient failure detection but may overlook critical visual cues and multimodal inputs, such as flowing water or audible sounds from a partially closed faucet, which are ignored in our current framework.
Thus, exploring more robust representations that balance real-time precision with minimal information loss or integrate richer multimodal inputs is a promising direction for future research.

\section{Constraint Painter}
\label{supsec: implementation details}
In this section, we first describe the data collection and annotation process of the Constraint-aware Segmentation Dataset. Next, we present the training details of the proposed {\mname} model. Finally, it is followed by a detailed explanation of the Element Pipeline.

\begin{figure*}[t]
\centering
\includegraphics[width=\linewidth]{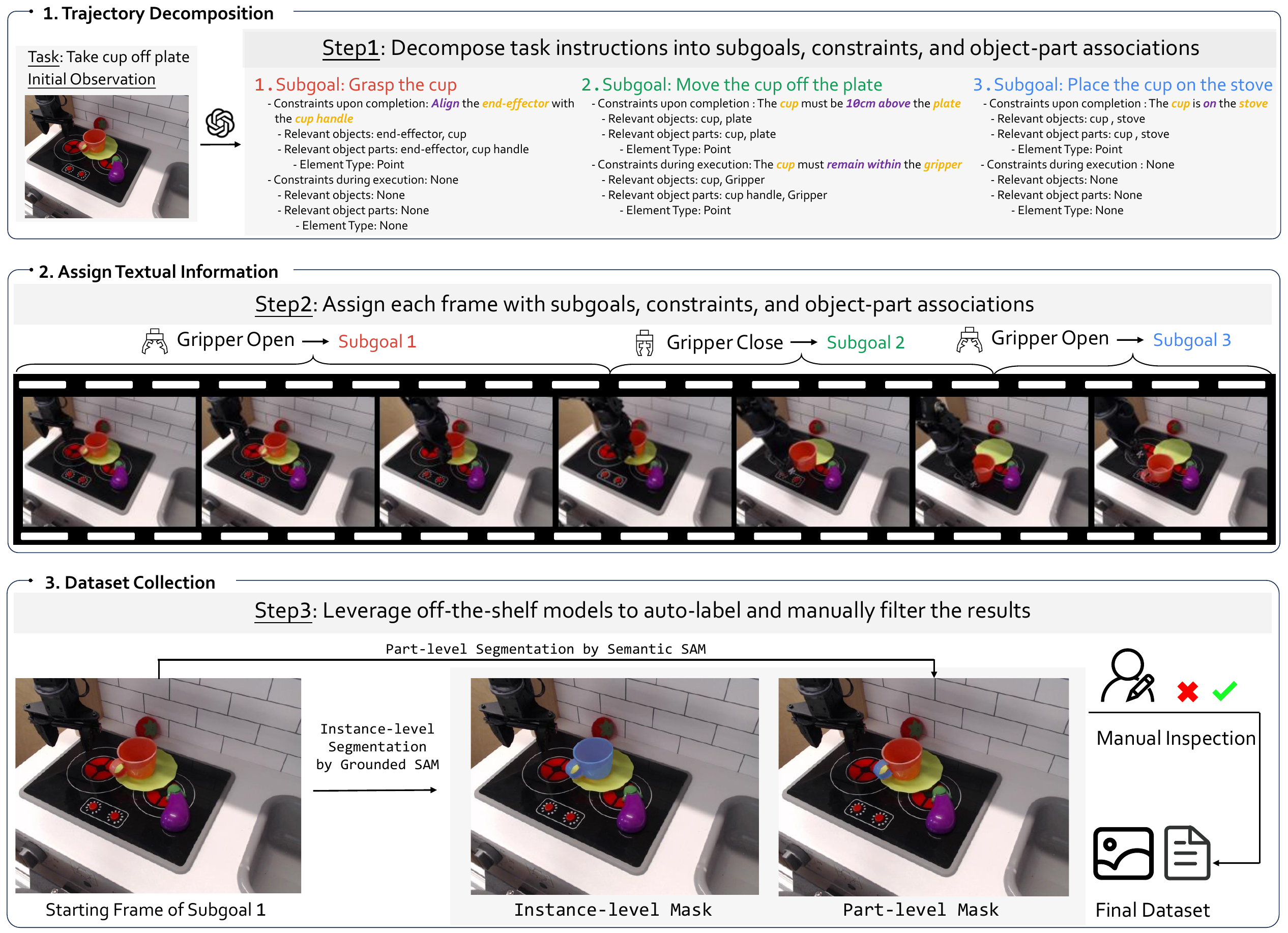}
   \caption{Dataset Collection Pipeline. Our data is sourced from BridgeData V2~\cite{li2022behavior}. The data collection process consists of three steps: \textbf{(1)} Using GPT-4o~\cite{achiam2023gpt} to decompose the task instruction based on the initial observation from the first frame of the trajectory, generating subgoals along with two types of constraints for each subgoal (\ie, constraints during execution and upon completion) and object-part associations. \textbf{(2)} Utilizing external references (\eg, gripper open/close states) to assign subgoals, constraints, and object-part associations to each frame. (3) Leveraging off-the-shelf models (\eg, Grounded SAM~\cite{ren2024grounded}, Semantic SAM~\cite{li2023semantic}) to generate instance- and part-level masks~(blue mask in this figure) automatically, followed by manual filtering to curate the final dataset. 
   }
\label{fig: sup data collection}
\end{figure*}

\subsection{{\mname} Data Collection}
To ensure broader coverage of scenarios and objects in our dataset, along with text-based instructions, we utilized the BridgeData V2 dataset~\cite{walke2023bridgedata}. This large and diverse dataset of robotic manipulation behaviors comprises $60,096$ trajectories collected across $24$ environments, encompassing $13$ distinct skills.

The dataset collection pipeline is shown in \cref{fig: sup data collection}. The entire process is divided into three stages, \ie, trajectory decomposition, assigning textual information, and dataset collection.
First, we decompose each trajectory’s instruction and initial observation from BridgeData V2 into subgoals for each stage, along with the constraints upon completion, constraints during execution, the corresponding object-part associations and element type for each constraint. 
Subfigure 1 in \cref{fig: sup data collection} illustrates a specific example, demonstrating the decomposition of the task ``Take cup off plate''.
Notably, the third subgoal, ``Place the cup on the stove'', indicates that the initial observation ensures the decomposition process fully understands the task's contextual environment.
During this process, we also obtain the constraint element type, which serves as the ground-truth text response for part-level constraint-aware segmentation.
This decomposition is performed using the off-the-shelf GPT-4o API.

After obtaining the subgoals, constraints, and object-part associations for each stage, we need to assign each frame to its corresponding stage.
% While BridgeData V2 provides trajectory-level instructions, we require frame-level annotations for per-frame subgoals and constraints.
%
Since BridgeData V2 does not provide per-frame annotations, we addressed this issue by sampling pick-and-place data and leveraging the additional information~(\eg, gripper open/close states) provided by BridgeData V2 for assignment. 
The pick-and-place task is typically divided into three stages: Approach, Grasp and Transfer, and Place, corresponding to the gripper states of open, closed, and open, respectively.
Leveraging the characteristics of the pick-and-place task, we complete the frame-level assignment. Subfigure 2 in \cref{fig: sup data collection} illustrates a specific example of the assignment process.

Using the obtained frame-level constraint-aware object and part information, Instance-level and part-level segmentations are performed using Grounded SAM~\cite{ren2024grounded} and Semantic SAM~\cite{li2023semantic}, respectively. 
We conducted a sampled manual inspection of the final annotations to filter out errors and low-quality labels. Our final dataset is composed of $10,181$ trajectories with $219,356$ images.
% using external references~(\eg, gripper open/close states) to generate $10,181$ trajectories with $219,356$ images. 
% %
% We employ GPT-4o to decompose trajectory instructions into subgoals, constraints, and object-part associations to generate ground-truth annotations. 
%

\begin{figure*}[t]
\centering
\includegraphics[width=\linewidth]{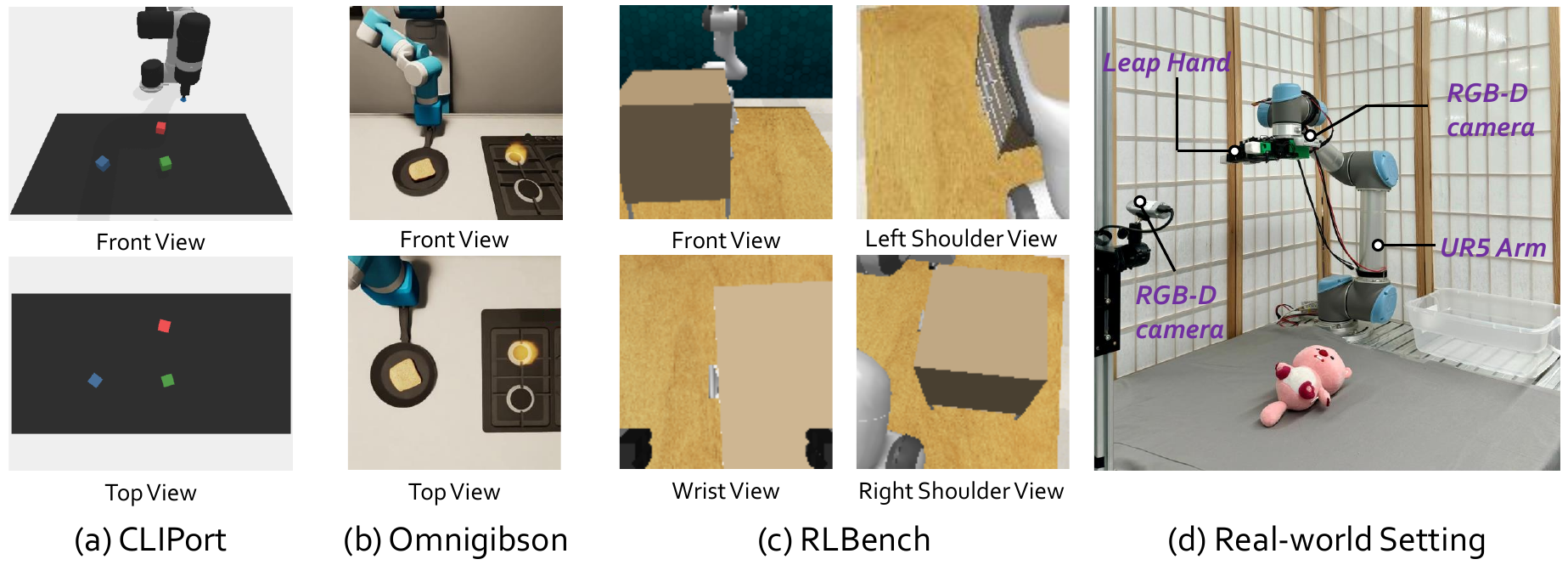}
   \caption{Environmental Setup of three simulators and one real-world setting.
   For CLIPort~\cite{shridhar2021cliport} and OmniGibson~\cite{li2022behavior}, we provide third-person front and top views and the robot platforms are the UR5 arm and Fetch, respectively. 
   RLBench~\cite{james2019rlbench} offers four camera views, including front left shoulder, right shoulder, and wrist views, with the robot platform being Franka equipped with a gripper. 
   We provide a wrist and a third-person front view for the real-world setting, utilizing a UR5 robot equipped with a Leap Hand~\cite{shaw2023leap}.
   }
\label{fig: sup obs space}
\end{figure*}

%
% These outputs are integrated and refined through manual inspection to produce the multi-granularity dataset, which is combined with LISA's training data to fine-tune our model. More details are in Supp.~A.
\subsection{{\mname} Training Details}
\noindent We adopt LISA's~\cite{lai2024lisa} loss function, including the next-token prediction loss for text output, and the combination of per-pixel BCE loss with DICE loss for mask output.
Our {\mname}-13B model is trained on an $8$ NVIDIA 80G H800 GPU for two days with a batch size of $4$.
Our training data comprises multiple components: SemanticSeg, ReferSeg, VQA, ReasonSeg, and ConstraintSeg, to ensure our model retains dialogue and reasoning segmentation capabilities while achieving constraint-aware segmentation.
LISA inspires this training data setting.
SemanticSeg includes ADE20K, COCO-Stuff, PACO-LVIS, and PASCALPart. ReferSeg includes refCLEF, refCOCO, refCOCO+, and refCOCOg. VQA includes LLaVAInstruct-150k. ConstraintSeg includes instance and part-level data.

% We collect a small dataset to fine-tune ConSeg for each simulator. 
The training setting described above is for the {\mname}-base model. Since the training data consists entirely of real-world scenarios, there is a significant gap between the simulation and real-world environments. 
To address this, the {\mname} model used in the simulation experiments is a fine-tuned version, called {\mname}-ft, finetuned on a small amount of data collected from the simulator. Specifically, we collect $100$ trajectories from each simulator, sampled frames at $1$ Hz, and utilize either ground truth masks from the simulator or annotations generated using Grounded SAM and Semantic SAM.
\highlight{Notably, we use the {\mname}-base model in real-world experiments, demonstrating the model's generalization capability across different scenarios.}

\subsection{Element Pipeline Details}
\label{supsubsec: element pipeline details}
Here, we provide additional details about the Constraint Element Pipeline. We first filter out outliers after obtaining the constraint-aware object/part point clouds. Next, we calculate the 3D spatial bounds occupied by the remaining point cloud and determine the voxel size for voxelization based on the element type.
We then perform point cloud clustering using the DBSCAN algorithm, which has advantages over other methods, including identifying clusters of arbitrary shapes, eliminating the need to predefine the number of clusters, and its effectiveness in high-density regions.

% Environment Setups, Policies, and Baselines

\section{Environment Configuration}
\label{supsec: environment configuration}

We first provide detailed descriptions of the simulators (CLIPort~\cite{shridhar2021cliport}, Omnigibson~\cite{li2022behavior}, RLBench~\cite{james2019rlbench}) and real-world setups used in our study. 
We then discuss the low-level control policies implemented in these environments. 
Finally, we present the baselines and their implementation specific to each environment.
\highlight{Notably, our framework, {\fname}, is policy agnostic, meaning it can be adapted to any control policy without requiring any modification.}

\subsection{Environmental Setup}

The CLIPort~\cite{shridhar2021cliport} simulator\footnote{\url{https://github.com/cliport/cliport}} is a robotic manipulation benchmark to gather extensive data for imitation learning and train a language-conditioned multi-task low-level control policy.
The environment features a UR5 robotic arm with a suction cup as the end effector for pick-and-place tasks and a spatula as the end effector for pushing tasks, both operating on a black tabletop. 
We use two cameras: a third-person front view and a top view to provide comprehensive perspectives of the tabletop, as shown in \cref{fig: sup obs space}~(a).

The OmniGibson~\cite{li2022behavior} simulator\footnote{\url{https://github.com/StanfordVL/OmniGibson}} offers a realistic setting including a physics engine capable of supporting features such as lighting rendering, gravity effects, and temperature variations impacting objects within the environment.
This platform also provides an extensive array of pre-configured scenes and objects, enabling researchers to customize setups and train mobile manipulation robots.
We select version 1.1.0 for our study.
This simulator involves a Fetch robot equipped with a gripper as the end effector, operating on a white tabletop. 
We utilize two cameras: a third-person front view and a top view to provide comprehensive perspectives of the tabletop, as shown in \cref{fig: sup obs space}~(b).

RLBench~\cite{james2019rlbench} simulator\footnote{\url{https://github.com/stepjam/RLBench}} is a widely used benchmark for robot manipulation, featuring tasks such as articulated objects and tool use. 
Researchers can gather data and train low-level control policies using imitation learning or reinforcement learning within this environment. 
RLBench features a Franka robotic arm with a gripper as its end effector, operating on a brown tabletop. Four cameras provide comprehensive tabletop coverage, as shown in \cref{fig: sup obs space}~(c).
Additionally, RLBench uses a sampling-based motion planner for motion planning given the next predicted action/pose.

In the real-world setup depicted in \cref{fig: sup obs space}~(d), we utilize a fixed UR5 robotic arm with a Leap Hand~\cite{shaw2023leap} as the end effector. 
Two RealSense D415 RGB-D cameras capture the scene, one mounted on the wrist and the other positioned for a third-person front view.

\begin{figure*}[t]
\centering
\includegraphics[width=\linewidth]{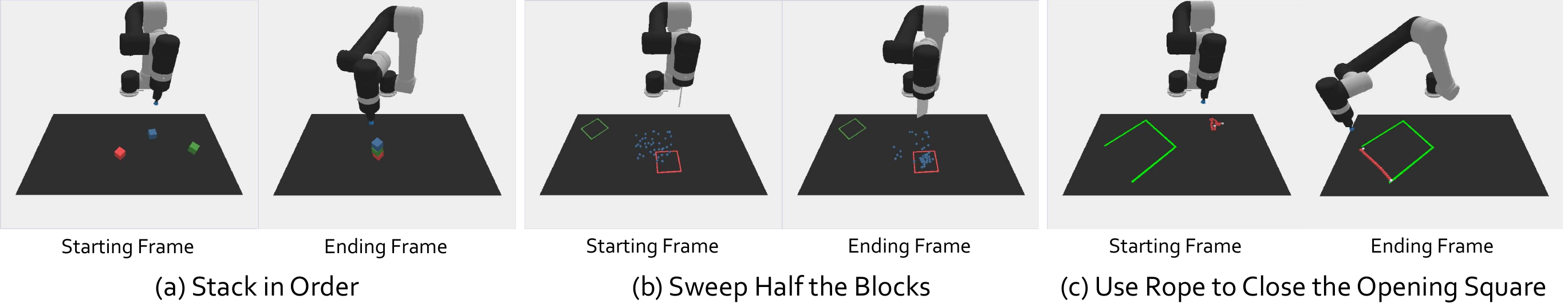}
   \caption{
   CLIPort task demonstration. we present three types of tasks in our experiments, including the starting and ending frames.
   }
\label{fig: sup cliport task}
\end{figure*}

\subsection{Control Policy}

In CLIPort, we use a pre-trained low-level policy the CLIPort~\cite{shridhar2021cliport} simulator provides to control the robotic arm and end effector. 
This policy can execute multi-task operations based on language instructions with RGB observations and its performance approaches perfection due to extensive imitation learning training. 
Notably, the policy is open-loop, meaning it does not adjust its actions in response to dynamic environmental changes~(\eg, it will not immediately pick up a dropped block during movement but will continue with the previously planned actions).

In Omnigibson, We utilize ReKep~\cite{huang2024rekep} as our low-level control policy, transforming long-horizon tasks into a set of relationships between fixed keypoints at different stages.
At each stage, an optimization algorithm computes these relationships to generate actions, enabling language-conditioned closed-loop control. 
Notably, ReKep employs a pre-trained large vision model (\ie, DINOv2~\cite{oquab2023dinov2}) to process raw RGB data, extracting semantically relevant keypoints. 
This approach also serves as the compared method in our ablation study for extracting 3D points through constraint-aware segmentation, showing our superiority.

In RLBench, we employ the Autoregressive Policy~(ARP)~\cite{zhang2024autoregressive} as the control policy, which generates the next action based on historical observations and action sequences through an autoregressive process. 
This method achieves state-of-the-art performance in the RLBench.

In the real-world setting, We employ DexGraspNet 2.0~\cite{zhang2024dexgraspnet} as our low-level policy, which predicts the dexterous hand's grasping pose based on the scene's point cloud and facilitates the robotic arm's action trajectory through motion planning to achieve robust generalized grasping.
{Notably, DexGraspNet 2.0 is an open-loop policy, which means it does not adjust to environmental changes during action execution.} 
For example, if the target object's position shifts while the arm moves, the system does not modify its motion plan. 
Therefore, it continues to execute toward the originally intended location to complete the grasp, failing.

\subsection{Baseline Details}

In CLIPort, we compare three baselines: CLIPort~\cite{shridhar2021cliport}, CLIPort with Inner Monologue~\cite{huang2022inner}, and CLIPort with DoReMi~\cite{guo2023doremi}. 
We slightly modify the original implementations of these three baselines to suit our task requirements.
\textbf{(1)} For CLIPort~\cite{shridhar2021cliport}, the sole modification involves substituting the original oracle success detector with an off-the-shelf VLM (\ie, GPT-4o~\cite{achiam2023gpt}) used as a failure detector. 
This change enables the robot system to determine whether to transition to the next subgoal using image-based vision question answering~(VQA). 
Notably, CLIPort decomposes the instructions into a list of subgoals before the task begins and does not dynamically adjust or revert to previous subgoals upon detecting a failure.
\textbf{(2)} For Inner Monologue~\cite{huang2022inner}, we replicate the implementation detailed in the original literature, by employing CLIPort for the low-level policy and an off-the-shelf VLM (\ie, GPT-4o~\cite{achiam2023gpt}) as the planner. 
This pipeline determines the next subgoal based on the completed subgoals and current observations after each subgoal concludes. 
Notably, Inner Monologue queries the VLM only at the end of each subgoal, without considering events that occur during execution.
\textbf{(3)} For DoReMi~\cite{guo2023doremi}, we reproduce the implementation according to the original DoReMi paper and enhance it by replacing the VLM initially~(\ie, BLIP2~\cite{li2023blip}) used for repeated VQA-style queries during robotic execution with a more powerful VLM (\ie, GPT-4o~\cite{achiam2023gpt}). 
Additionally, we substitute its LLM, which lacks environmental awareness, with the same GPT-4o to serve as the task planner.

\begin{table*}[t]
\caption{Detailed Performance in CLIPort. We report the success rate and execution time for three tasks, compared to baseline methods.}
\centering
\small
\renewcommand{\arraystretch}{1.2} % 增加行间距
\setlength{\tabcolsep}{1.5pt}
\begin{tabular}{lc|c|ccc|ccc}
\bottomrule[1pt]
\multicolumn{2}{c|}{Tasks with}                  & \multicolumn{4}{c}{Success Rate(\%) $\uparrow$} & \multicolumn{3}{c}{Execution Time(s) $\downarrow$} \\
\multicolumn{2}{c|}{disturbance}                & CLIPort & +Inner Monologue  & +DoReMi & +\textbf{Ours}  & +Inner Monologue & +DoReMi & +\textbf{Ours} \\
\hline
  Stack in         & $p$=0.0                   & 100.00 $\pm$ 0.00 & 100.00 $\pm$ 0.00  & 100.00 $\pm$ 0.00 & 100.00 $\pm$ 0.00    & 13.40 $\pm$ 1.82 & 13.40 $\pm$ 1.82 & 13.40 $\pm$ 1.82   \\
  order with       & $p$=0.15                  & 56.67 $\pm$ 6.11  & 81.67 $\pm$ 6.11  & 83.33 $\pm$ 5.17 & \textbf{95.00 $\pm$ 4.00}        & 34.80 $\pm$ 3.12 & 26.00 $\pm$ 2.77 & \textbf{21.00 $\pm$ 1.75} \\
  drop $p$         & $p$=0.3                   & 21.67 $\pm$ 8.33 & 75.00 $\pm$ 8.95  & 76.67 $\pm$ 9.52 & \textbf{88.33 $\pm$ 6.53}    & 42.80 $\pm$ 3.18 & 34.20 $\pm$ 2.73 & \textbf{25.40 $\pm$ 2.95} \\
\hline
  Stack in         & $q$=1                     & 90.00 $\pm$ 6.11  & 90.00 $\pm$ 6.11  & 96.67 $\pm$ 4.00 & \textbf{98.33 $\pm$ 3.27}    & 24.80 $\pm$ 4.08 & 24.60 $\pm$ 4.66 & \textbf{24.20 $\pm$ 4.65}   \\
  order with       & $q$=2                     & 41.67 $\pm$ 7.30 & 71.67 $\pm$ 8.33  & 75.00 $\pm$ 5.17 & \textbf{83.33 $\pm$ 5.17}        & 39.40 $\pm$ 5.87 & 37.00 $\pm$ 6.29 & \textbf{29.20 $\pm$ 4.61} \\
  noise $q$        & $q$=3                     & 15.00 $\pm$ 6.11 & 40.00 $\pm$ 8.00  & 40.00 $\pm$ 6.11 & \textbf{63.33 $\pm$ 8.33}    & 58.20 $\pm$ 4.74 & 54.20 $\pm$ 6.02 & \textbf{36.80 $\pm$ 4.61} \\
\hline
  \multicolumn{2}{c|}{Sweep Half the Blocks}   & 0.00 $\pm$ 0.00 & 18.33 $\pm$ 6.11  & 16.67 $\pm$ 8.95 & \textbf{75.00 $\pm$ 11.55}    & 22.00 $\pm$ 2.91 & 16.60 $\pm$ 1.33 & \textbf{16.40 $\pm$ 1.00}   \\
\hline
  \multicolumn{2}{l|}{Use Rope to Close}   & \multirow{2}{*}{0.00 $\pm$ 0.00} & \multirow{2}{*}{68.33 $\pm$ 9.52}  & \multirow{2}{*}{58.33 $\pm$ 18.62} & \multirow{2}{*}{\textbf{76.67 $\pm$ 6.11}}    & \multirow{2}{*}{41.60 $\pm$ 6.34} & \multirow{2}{*}{65.80 $\pm$ 7.40} & \multirow{2}{*}{\textbf{34.60 $\pm$ 2.81}}   \\
  \multicolumn{2}{l|}{the Opening Square}   &  &   &  &     &  &  &    \\

\bottomrule[1pt]
\end{tabular}
\label{tab: sup cliport}
\end{table*}

In Omnigibson, we compare two baselines: ReKep~\cite{huang2024rekep} and ReKep with DoReMi~\cite{guo2023doremi}. 
\textbf{(1)} For ReKep, we directly implement it using the official codebase. 
\textbf{(2)} For DoReMi, it is implemented as described above.

In RLBench, we compare two baselines: ReKep~\cite{huang2024rekep} and ReKep with DoReMi~\cite{guo2023doremi}. 
\textbf{(1)} For ReKep, we directly implement it using the official codebase. 
\textbf{(2)} For DoReMi, it is implemented as described above.

For the real-world evaluation, we compare two baselines: 
DexGraspNet 2.0~\cite{zhang2024dexgraspnet} and DexGraspNet 2.0 with DoReMi~\cite{guo2023doremi}. 
\textbf{(1)} For DexGraspNet 2.0, we directly implement it using the official codebase. 
\textbf{(2)} For DoReMi, it is implemented as described above.

\section{Evaluation Details}
\label{supsec: evaluation details}

In this section, we first detail the task specifications within the simulator and real-world evaluations.
Then, we introduce the disturbances introduced in each task and the evaluation metrics used. 
Finally, we report the detailed experimental results and our analyses, with additional results not included in the main text due to space constraints.

\subsection{CLIPort}
\label{supsubsec: cliport evaluation details}

\subsubsection{Task, Disturbance and Metric Details}

As shown in \cref{fig: sup cliport task}, we evaluate three tasks in CLIPort:
\textbf{(1)} ``Stack in Order'': Given blocks in red, green, and blue on a table, the robot must stack them with red at the bottom, green in the middle, and blue on top.
\textbf{(2)} ``Sweep Half the Blocks'' With $40$ blocks on the table, the robot must sweep approximately half of them (with a permissible error margin of $\pm 10\%$, \ie, $16$ to $24$ blocks) to a designated area.
\textbf{(3)} ``Use Rope to Close the Opening Square'': The robot should use a rope to enclose an open square, to enclose the area sufficiently, rather than form a perfectly closed square.

\begin{figure*}[t]
\centering
\includegraphics[width=\linewidth]{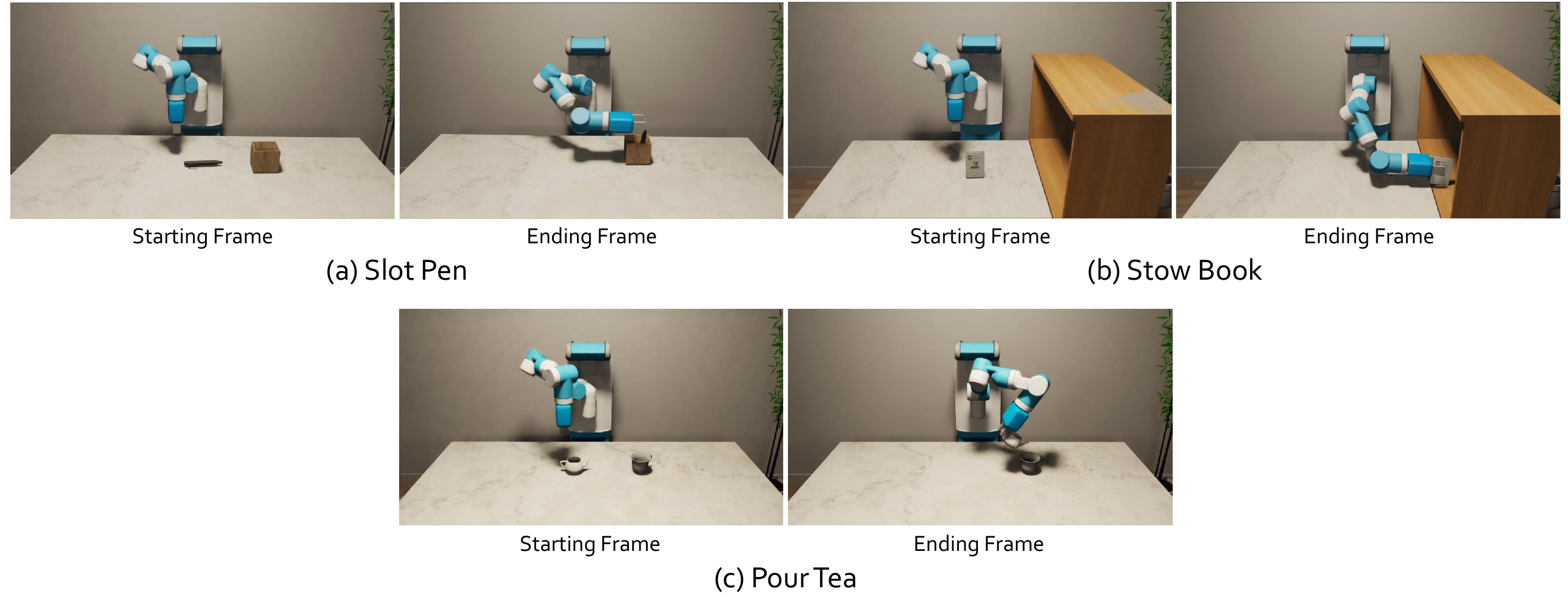}
   \caption{
   Omnigibson task demonstration. we present three types of tasks in our experiments, including the starting and ending frames.
   }
\label{fig: sup omnigibson task}
\end{figure*}

We introduce two types of disturbances to the ``Stack in Order'' task: 
\textbf{(1)} after the suction cup grasps a block, there is a probability $p$ at each step that the block will be released, causing it to drop; 
\textbf{(2)} The predicted placement position by the policy is perturbed by a uniform noise in the range $[0,q]$ cm, potentially leading to block tower collapse.

For each task and disturbance type, we conduct $5$ trials using different seeds, each comprising $12$ episodes.
We assess performance based on success rate and execution time, \highlight{excluding} the computational time for invoking the VLM. 
Results are reported as mean values with $95\%$ confidence intervals.
In the ``Stack in Order'' task, the robot must successfully stack the blocks in the specified order into a tower within $70$ seconds, despite the two perturbations above.
For the ``Sweep Half the Blocks'' task, the pre-trained policy aims to sweep blocks into a designated area. The robot must stop the policy once half of the blocks are in the target region. If, after $30$ seconds, the number of blocks in the area falls within the required range ($16$–$24$), the task is considered successful.
For the ``Use Rope to Close the Opening Square'' task, the pre-trained policy attempts to close an open rectangle into a perfect square using a rope. The robot must detect when the rectangle is sufficiently enclosed and immediately stop execution. Success is achieved if the robot halts within $70$ seconds, and the enclosure is complete.

\subsubsection{Detailed Experiment Results}
In \cref{tab: sup cliport}, we present detailed results in CLIPort, including those discussed in the main text and additional results.

In the ``Stack in Order'' task under severe interference conditions, our {\fname} shows an improvement of $18.33\%$ and $17.5\%$ in success rate over Inner Monologue~\cite{huang2022inner} and DoReMi~\cite{guo2023doremi}, respectively, while also reducing execution times by $38.7\%$ and $14.4\%$ compared to Inner Monologue and DoReMi, respectively.
%
% We find that utilizing code with elements enables both \textit{reactive} and \textit{proactive} failure detection, allowing for real-time identification and prevention of failure. 
%
% Furthermore, calculating the spatial relationships of elements through code provides superior monitoring of 3D spatial relations compared to direct image query via VLM. 
%
% These findings are discussed in the main text, and the processes of failure detection and recovery are shown in \cref{xxx}.
%
The failure detection and recovery processes are shown in \cref{fig: sup cliport demo 1} and \cref{fig: sup cliport demo 2}.

In the ``Sweep Half the Blocks'' task, our {\fname} achieved success rates that are $4.1$ and $4.5$ times higher than those of Inner Monologue~\cite{huang2022inner} and DoReMi~\cite{guo2023doremi}, respectively. 
%
% This means that directly computing the 3D positions of block-representative points to verify their presence within a rectangular area offers greater precision than relying on VLMs for image interpretation, as outlined in the main text.  
%
However, the success rate is not very high even in distraction-free scenarios.  
This is attributed to the high density of tracking points in the scene, which increases the likelihood of confusion and tracking errors, leading to inconsistent block counts within the target area.
The completion process of the task is also illustrated in \cref{fig: sup cliport demo 3}.

In the ``Use Rope to Close the Opening Square'' task, our approach outperforms Inner Monologue~\cite{huang2022inner} and DoReMi~\cite{guo2023doremi} by $8.34\%$ and $18.34\%$ in success rates, respectively, while also reducing execution times by $16.82\%$ and $47.43\%$ compared to Inner Monologue and DoReMi, respectively.
We find that calculating the distance between the rope ends and the opening's edges to determine closure is more accurate than directly querying a VLM with an image, allowing for earlier termination of the policy execution. 
The complete processes of the task are illustrated in \cref{fig: sup cliport demo 4}.

\subsection{OmniGibson}
\label{supsubsec: omnigibson evaluation details}

\subsubsection{Task, Disturbance and Metric Details}

As shown in \cref{fig: sup omnigibson task}, in the OmniGibson environment, we evaluated three distinct tasks:
\textbf{(1)} Slot Pen: A pen placed on a desk is picked up, rotated to a near-vertical position, moved above a pen holder, and then inserted into the holder.
\textbf{(2)} Stow Book: A book located on a desk is picked up and vertically positioned on a bookshelf.
\textbf{(3)} Pour Tea: A teapot on the desk is lifted, horizontally moved above a teacup, and then tilted to pour tea into the cup.

We introduce three types of disturbances with varying constraint elements for each task:
\textbf{(1)} Slot Pen Task: Point-level disturbances are applied as follows: (a) moving the pen while the robot is grasping it, (b) forcing the robot to release the pen mid-transfer, causing it to drop onto the table, and (c) moving the pen holder while the robot attempts to insert the pen. Despite these disturbances, the task is considered successful only if the robot can insert the pen into the holder.
\textbf{(2)} Stow Book Task: Line-level disturbances include: (a) rotating the book during the robot’s grasping process, (b) altering the book’s pose during transfer to disrupt its vertical alignment, and (c) reorienting the book horizontally after it has been placed vertically on the shelf. Success requires the robot to place the book vertically on the shelf despite these disturbances.
\textbf{(3)} Pour Tea Task: Surface-level disturbances involve: (a) tilting the container forward or backward during transfer, (b) inducing lateral tilts during movement, and (c) restoring the container to a level position during pouring. To succeed, the robot must prevent spillage and complete the pouring task under these disturbances.

We conduct experiments on three tasks, each consisting of one no-disturbance trial and three specific-disturbance trials. Each trial was repeated 10 times, and the performance was evaluated based on success rate, execution time (\highlight{including} the computation time for invoking the VLM), and the number of tokens used.

\subsubsection{Detailed Experiment Results}

Specific experimental results are detailed in the main text; here, we present additional demonstrations in Sec.\ref{supsubsec: omniGibson demonstrations}.

\begin{figure*}[t]
\centering
\includegraphics[width=\linewidth]{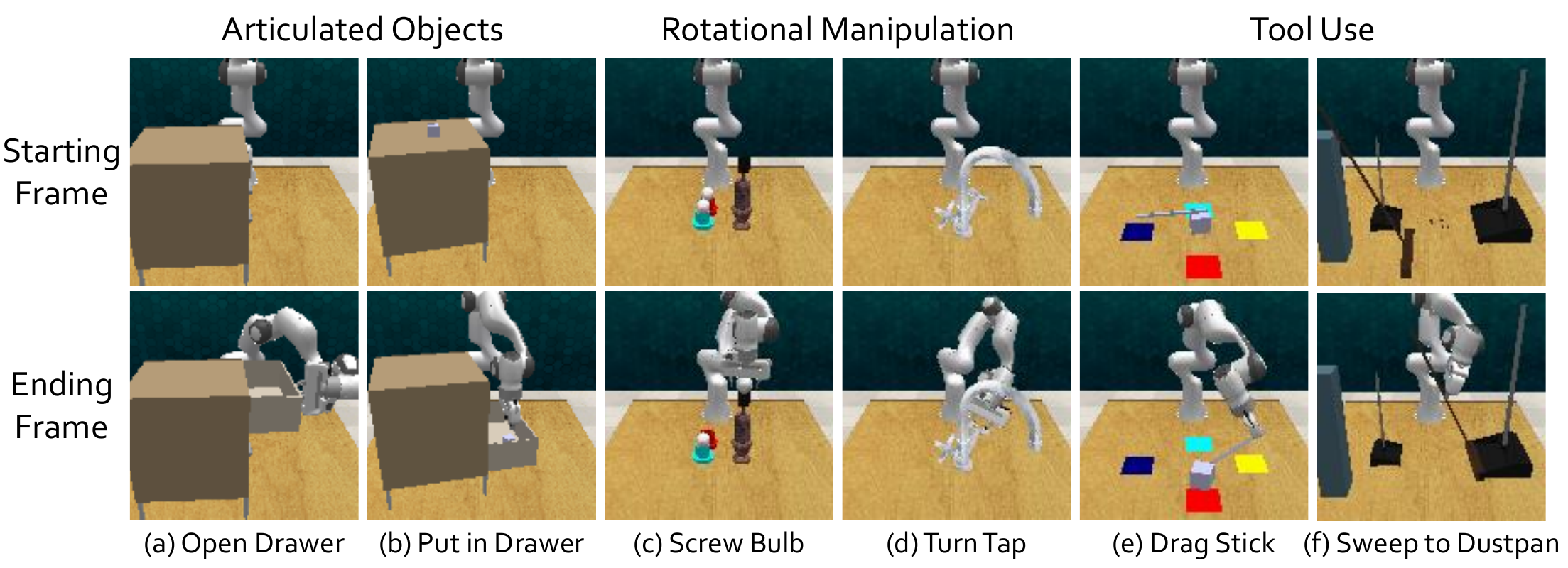}
   \caption{
   RLBench task demonstration. we present six types of tasks in our experiments, including the starting and ending frames.
   }
\label{fig: sup rlbench task}
\end{figure*}

\begin{table*}[t]
\caption{Performance in RLBench. We report the success rate, compared to baseline methods.}
\small
\centering
\begin{tabular}{l|c|cc|cc|cc}
\bottomrule[1pt]
\multirow{2}{*}{Method}        &   Avg.                     
                               & \multicolumn{2}{c|}{Articulated Object} 
                               & \multicolumn{2}{c|}{Tool-Use} 
                               & \multicolumn{2}{c}{Tool-Use}  
                               \\
\cline{3-8}
                               &  Success  Rate~(\%) $\uparrow$                 &  Open Drawer    &   Put in  Drawer                   & Screw Bulb  & Turn  Tap               &  Drag Stick     &   Sweep to  Dustpan      \\
\hline

 RVT2~\cite{goyal2024rvt2}          &89.83         &90.3     &97.6    &86.6   &91.0   &93.8     &79.7 \\  
ARP~\cite{zhang2024autoregressive}  &91.27         &93.9     &91.0    &86.4   &96.6   &88.1     &91.6 \\        

\hline

~~+DRM~\cite{guo2023doremi}         &87.97         &90.6     &87.7   &83.1   &93.3    &84.8     &88.3 \\               
~~\textbf{+Ours}                    &\textbf{97.08}         & \textbf{98.1}     &\textbf{98.3}   &\textbf{97.5}   &\textbf{97.9}     &\textbf{95.6}     &\textbf{94.0} \\                          

\bottomrule[1pt]
\end{tabular}
\label{tab: rlbench}
\end{table*}

\subsection{RLBench}
\label{supsubsec: rlbench evaluation details}

\subsubsection{Task, Disturbance and Metric Details}

As shown in \cref{fig: sup rlbench task}, in RLBench, we evaluate six tasks across three categories of manipulation:
\textbf{(1)} Articulated Object Interaction: (a) Open Drawer——Open the top drawer (b) Put in Drawer——Open the drawer and place an item into the open drawer.
\textbf{(2)} Rotational Manipulation: (a) Screw Bulb——Screw in the red light bulb. (b) Turn Tap——Turn the left tap.
\textbf{(3)} Tool Use: (a) Drag Stick——Use a stick to drag the cube onto the red target. (b) Sweep to Dustpan——Sweep dirt into the tall dustpan.

In RLBench, we avoid introducing additional disturbances, as its control policy naturally generates diverse failures to validate the effectiveness of our framework.
The RLBench-trained policy lacks failure recovery mechanisms; thus, any episode flagged as a failure by the detection framework is deemed invalid, and a new episode is initiated.

For each task, we evaluate performance over $1,000$ valid episodes (maximum $25$ steps each), measured by the average success rate.

\subsubsection{Detailed Experiment Results}

% 
% 值得注意的是，尽管我们的训练数据中没有任何有关铰链物体的数据，
\highlight{Code with elements can generalize better to monitor diverse tasks.}
\highlight{Notably, despite our training data lacking information on articulated objects at both the instance and part levels, our method effectively handles them, accurately segmenting parts such as drawer handles}. 
In the ``Open Drawer'' task, {\fname} achieves a $98.1\%$ success rate, significantly outperforming DRM's $90.6\%$.
For ``Put in Drawer'', {\fname} reaches $98.3\%$, surpassing DRM's $87.7\%$ by $10.6$ percentage points. 
\highlight{We attribute this generalization to two factors: \textbf{(1)} the inherent prior world knowledge of extensively pre-trained VLMs (\eg, SAM, LLaVA), which enables generalization to unseen tasks; and \textbf{(2)} our minimalist scheme that abstracts articulated objects into geometric components via constraint elements, ignoring irrelevant details, enhancing the generalizability.}

We also demonstrate our method on additional unseen tasks, such as rotational manipulation and tool use, where it consistently outperforms baseline methods.

\subsection{Real-world Evaluation}
\label{supsubsec: real-world evaluation details}

\begin{table*}[t]
\caption{Detailed Performance of Single Pick \& Place. We report the success rate and execution time. DGN is DexGraspNet 2.0~\cite{zhang2024dexgraspnet}.}
\centering
\small
\begin{tabular}{l|cc|c|cc|cc}
\bottomrule[1pt]
Tasks with         & Object   & Object  & \multicolumn{3}{c}{Success Rate(\%) $\uparrow$} & \multicolumn{2}{c}{Execution Time(s) $\downarrow$} \\
disturbance        & types    & Name    & DGN  & +DoReMi & +\textbf{Ours}   & +DoReMi & +\textbf{Ours} \\

\hline
 Pick \& Place with         & \multirow{3}{*}{Deformable}   & Toy Loopy   & 0.00  & 80.00 & \textbf{100.00}        & 64.91 $\pm$ 2.83 & \textbf{46.02$\pm$  3.11}   \\
 the objects being          &    & Toy Dog   & 0.00  & 80.00 & \textbf{100.00}        & 60.68 $\pm$ 4.00 & \textbf{47.06 $\pm$ 3.24}   \\
 moved during               &    & Toy Rabbit   & 0.00  & 90.00 & \textbf{90.00}        & 59.83 $\pm$ 1.82 & \textbf{45.77 $\pm$ 2.03}   \\
                            \cline{2-8}
 grasping                   & \multirow{3}{*}{Transparent}  & Beverage Bottle   & 0.00  & 60.00 & \textbf{100.00}        & 69.97 $\pm$ 7.89 & \textbf{47.61 $\pm$ 2.58} \\
                            &   & Glass Cup   & 0.00  & 70.00 & \textbf{90.00}        & 76.99 $\pm$ 4.60 & \textbf{48.32 $\pm$ 3.22} \\
                            &   & Shampoo Bottle   & 0.00  & 70.00 & \textbf{90.00}        & 70.91 $\pm$ 5.68 & \textbf{48.31 $\pm$ 3.08} \\
                            \cline{2-8}
                            & \multirow{3}{*}{Small Rigid}  & Apple Model   & 0.00  & 80.00 & \textbf{100.00}        & 64.65 $\pm$ 4.34 & \textbf{45.39 $\pm$ 0.71} \\
                            &   & Pear Model   & 0.00  & 90.00 & \textbf{90.00}         & 67.11 $\pm$ 1.10 & \textbf{45.48 $\pm$ 1.01} \\
                            &   & Peach Model   & 0.00  & 70.00 & \textbf{90.00}        & 65.48 $\pm$ 2.90 & \textbf{45.37 $\pm$ 0.64} \\
                            \cline{2-8}
                            & \multirow{3}{*}{Large Geometric} & Plate & 0.00  & 80.00 & \textbf{100.00}        & 69.86 $\pm$ 2.64 & \textbf{45.18 $\pm$ 0.65} \\
                            &  & Ball & 0.00  & 90.00 & \textbf{100.00}        & 67.43 $\pm$ 2.63 & \textbf{45.37 $\pm$ 0.70} \\
                            &  & Pyramid & 0.00  & 90.00 & \textbf{90.00}        & 69.14 $\pm$ 3.32 & \textbf{45.42 $\pm$ 0.72} \\
\hline
 Pick \& Place with         & \multirow{3}{*}{Deformable}   & Toy Loopy   & 0.00  & 80.00 & \textbf{90.00}        & 69.29 $\pm$ 4.87 & \textbf{60.86 $\pm$ 3.41}   \\
 the objects being          &    & Toy Dog   & 0.00  & 70.00 & \textbf{100.00}        & 66.09 $\pm$ 2.99 & \textbf{63.12 $\pm$ 3.75}   \\
 removed during             &    & Toy Rabbit   & 0.00  & 80.00 & \textbf{90.00}        & 70.86 $\pm$ 4.56 & \textbf{63.40 $\pm$ 3.88}   \\
                            \cline{2-8}
 movement                   & \multirow{3}{*}{Transparent}  & Beverage Bottle   & 0.00  & 50.00 & \textbf{90.00}        & 77.90 $\pm$ 2.89 & \textbf{61.97$\pm$  3.90} \\
                            &   & Glass Cup   & 0.00  & 70.00 & \textbf{90.00}        & 70.00 $\pm$ 3.46 & \textbf{63.22 $\pm$ 4.35} \\
                            &   & Shampoo Bottle   & 0.00  & 60.00 & \textbf{90.00}        & 60 $\pm$ 4.28 & \textbf{63.00 $\pm$ 3.81} \\
                            \cline{2-8}
                            & \multirow{3}{*}{Small Rigid}  & Apple Model   & 0.00  & 70.00 & \textbf{90.00}        & 70.21 $\pm$ 4.30 & \textbf{63.71 $\pm$ 3.91} \\
                            &   & Pear Model   & 0.00  & 60.00 & \textbf{100.00}        & 72.70 $\pm$ 4.84 & \textbf{58.61$\pm$  2.41} \\
                            &   & Peach Model   & 0.00  & 60.00 & \textbf{90.00}        & 66.48 $\pm$ 3.32 & \textbf{59.19 $\pm$ 2.59} \\
                            \cline{2-8}
                            & \multirow{3}{*}{Large Geometric} & Plate & 0.00  & 90.00 & \textbf{100.00}        & 72.00 $\pm$ 2.77 & \textbf{59.21 $\pm$ 2.61} \\
                            &  & Ball & 0.00  & 70.00 & \textbf{100.00}        & 70.92 $\pm$ 3.37 & \textbf{61.57 $\pm$ 3.80} \\
                            &  & Pyramid & 0.00  & 70.00 & \textbf{90.00}        & 73.83 $\pm$ 2.82 & \textbf{60.25 $\pm$ 3.25} \\
\bottomrule[1pt]
\end{tabular}
\label{tab: sup hand 1}
\end{table*}

\subsubsection{Task, Disturbance and Metric Details}
We conduct real-world evaluations on two tasks:
\textbf{(1)} Simple Pick \& Place: The robot should pick and place objects within $70$ seconds. 
We include four kinds of objects: Deformable, Transparent, Small Rigid, and Large Geometric, with three objects in each category. 
The deformable objects are Loopy,  Dog, and  Rabbit toys, which undergo deformation when grasped by the dexterous hand.
The transparent objects are a clear beverage bottle, a transparent glass, and a clear shampoo bottle. 
The small rigid objects consist of apple, pear, and peach models, while the large geometric objects include a large plate, ball, and pyramid.
\textbf{(2)} Reasoning Pick \& Place: In cluttered scenes, the robot must perform long-horizon tasks where the instructions contain ambiguous terms (\eg, ``animals'' or ``fruits'' without specifying particular types). 
Specifically, the tasks are: (a) ``Clear all objects on the table except for \underline{animals}'', and (b) "Grasp the \underline{animals} according to their distances to \underline{fruits}, from nearest to farthest'', with ambiguous terms underlined.

For both tasks, we introduce two identical disturbances:
\textbf{(1)} Moving the object during the robot's grasping.
\textbf{(2)} Removing the object from the robot's dexterous hand during movement after grasping.

For each task, and each object involved in Simple Pick \& Place, we conduct 10 trials.
For each long-horizon task in Reasoning Pick \& Place, we also conduct 10 trials.
We evaluate performance based on success rate and execution time, \highlight{including} the computational time invoking the VLM.
Results are reported as mean values with $95\%$ confidence intervals.
For the Simple Pick \& Place, the robot has only one opportunity to autonomously release the object held in its gripper at a designated location. 
Any disturbance the robot encounters allows for a return and reattempt at grasping if the robot successfully detects it. 
Success is defined as meeting these conditions within $90$ seconds.
For the Reasoning Pick \& Place task, the robot must clear all objects on the table except for animals within $4$ minutes for ``Clear all objects on the table except for animals''. 
In the task ``Grasp the animals according to their distances to fruits, from nearest to farthest'', the robot must sequentially grasp the animals in the correct order within $2$ minutes, despite human-induced distractions such as moving animals or fruits.
\highlight{Notably, this task is particularly challenging because the robot operates under an open-loop policy, preventing it from using closed-loop feedback to handle the dynamic distances between fruits and animals caused by external disturbances.
Therefore, a failure detection framework is necessary to enable both \textit{reactive} and \textit{proactive} real-time detection with high precision, monitoring the distance changes and adjusting the grasping sequence accordingly.}

\subsubsection{Detailed Experiment Results}

In \cref{tab: sup hand 1}, we present detailed results of Simple Pick \& Place.
{\fname} achieves success rates surpassing DoReMi by $20.4\%$ when handling different kinds of objects.
We show real-world demonstrations of Simple Pick \& Place and Reasoning Pick \& Place in Sec.~\ref{supsubsec: real-world demonstrations}.

\section{More ablation studies}
\label{supsec: more ablation}

\textbf{Segmentation model ablations.}
\cref{tab: rebuttal omnigibson} presents further ablation studies, replacing {\mname} with LISA and PixelLM.  
Our {\mname} shows superior overall framework performance (check \cref{tab: rebuttal omnigibson} ID A \& B \& E), which represents a key technical contribution.
%
% Tab.~5 in the main paper also evaluates these models using segmentation metrics.

\vspace{+1mm}
\noindent\textbf{Failure detection mode ablations.}
% \cref{tab: rebuttal omnigibson} provides additional ablation results:
%
\textbf{(1)} In \cref{tab: rebuttal omnigibson}~(ID C), proactive failure detection alone yields a lower success rate due to its inability to handle unforeseen failures;
\textbf{(2)} In \cref{tab: rebuttal omnigibson}~(ID D), reactive detection alone achieves a slightly higher success rate but incurs longer execution times, as it only responds post-failure;
\textbf{(3)} In \cref{tab: rebuttal omnigibson}~(ID E), the synergy of both modes achieves the best performance by addressing the limitations of each mode.

\section{More Demonstrations and Prompts}
\label{supsec: more demonstrations and prompts}

This section presents additional demonstrations, including simulations and real-world scenarios of failure detection and recovery, as well as constraint-aware segmentation results and prompts.

\subsection{CLIPort}
\label{supsubsec: cliport demonstrations}

% 这里我们展示了 CLIPort 中三个任务 ``Stack in Order'' ``Sweep Half the Blocks'' ``Use Rope to Close the Opening Square'' 的 demonstrations

Here, we present demonstrations of three tasks in CLIPort: ``Stack in Order'', ``Sweep Half the Blocks'', and ``Use Rope to Close the Opening Square''.

% \cref{fig: sup cliport demo 1} 展示了 ``Stack in Order'' 这个task 面临policy预测的放置位置会受到uniform [0,q] cm 的干扰的情况下，我们的框架如何 failure detection 和并辅助 recovery的。

\begin{figure*}[t]
\centering
\includegraphics[width=\linewidth]{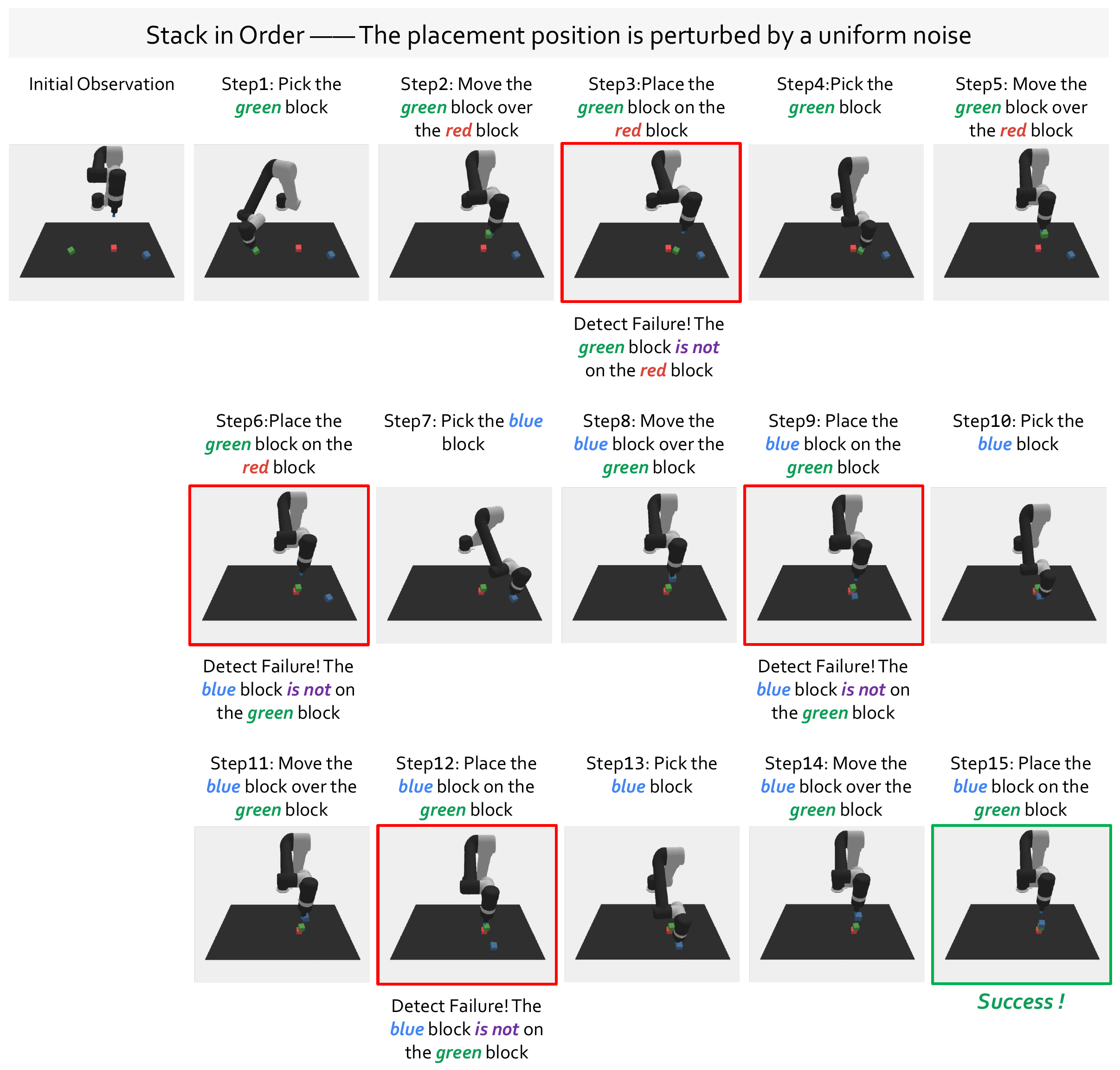}
   \caption{
   Demonstration of ``Stack in Order''. We show how our framework detects failures and assists in recovery when the placement positions predicted by the policy for the ``Stack in Order'' task are subject to a uniform $[0,q]$ cm interference. Red boxes indicate the occurrence of failures, while green boxes signify successful task execution.
   }
\label{fig: sup cliport demo 1}
\end{figure*}

\cref{fig: sup cliport demo 1} demonstrates how our framework detects failures and assists in recovery when the placement positions predicted by the policy for the ``Stack in Order'' task are subject to a uniform $[0,q]$ cm interference.

% \cref{fig: sup cliport demo 2} 展示了 ``Stack in Order'' 这个task 面临每一步都有p的概率会方块被机器人的吸盘释放而掉落的情况下，我们的框架如何 failure detection 和并辅助 recovery的。

\begin{figure*}[t]
\centering
\includegraphics[width=\linewidth]{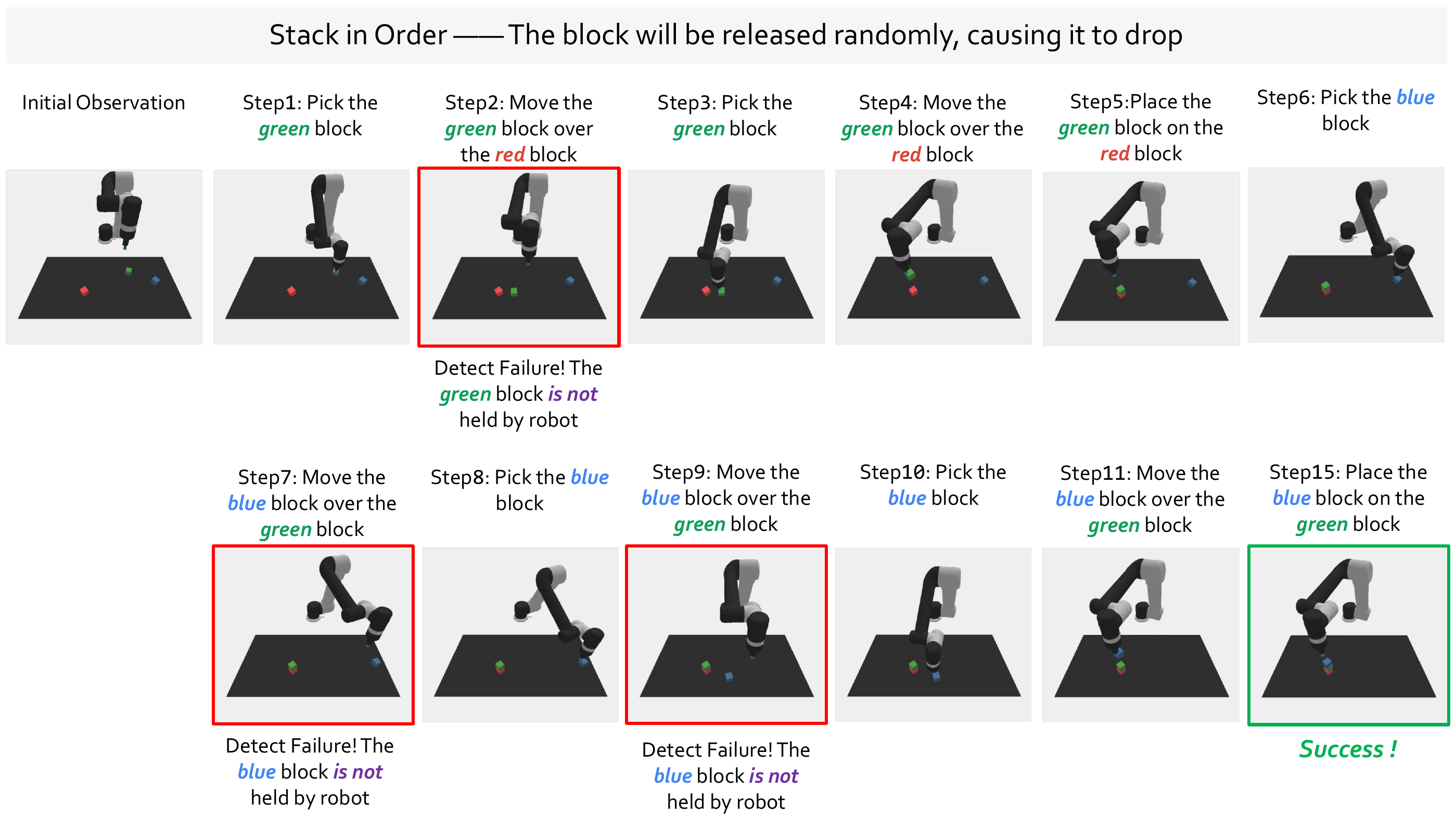}
   \caption{
   Demonstration of ``Stack in Order''. We show how our framework performs failure detection and aids in recovery when, in the ``Stack in Order'' task, there is a probability \( p \) that blocks will fall due to being released by the robot's suction cup at each step. Red boxes indicate the occurrence of failures, while green boxes signify successful task execution.
   }
\label{fig: sup cliport demo 2}
\end{figure*}

\cref{fig: sup cliport demo 2} illustrates how our framework performs failure detection and aids in recovery when, in the ``Stack in Order'' task, there is a probability \( p \) that blocks will fall due to being released by the robot's suction cup at each step.

% \cref{fig: sup cliport demo 3} 展示了 ``Sweep Half the Blocks'' 这个task，我们框架可以快速对指定区域内的方块计数，并及时停止policy执行以成功完成任务，而DoReMi无法及时停止policy执行，导致任务失败。

\begin{figure*}[t]
\centering
\includegraphics[width=\linewidth]{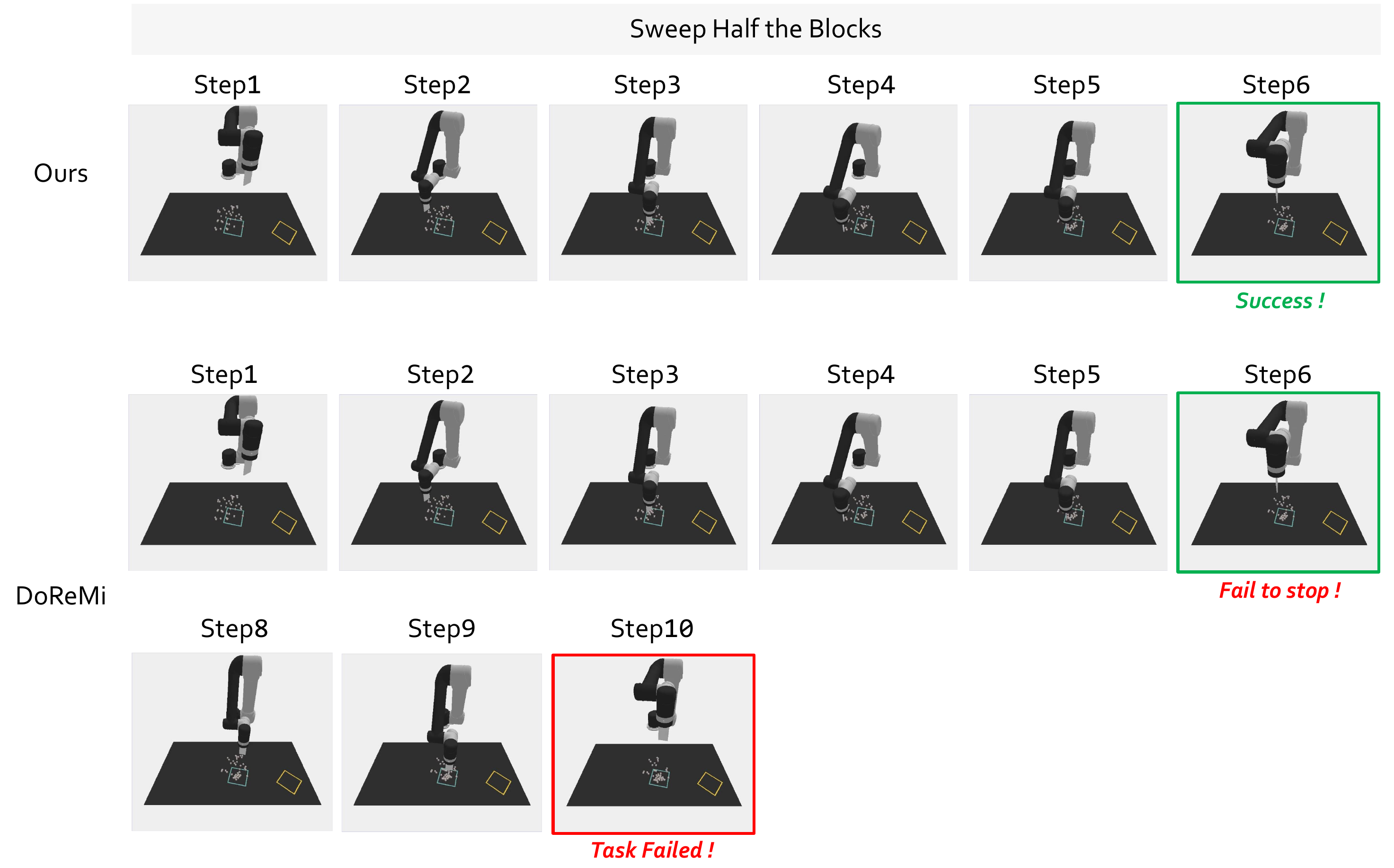}
   \caption{
   Demonstration of ``Sweep Half the Blocks'' and comparison to baseline. We show our framework can precisely count the blocks within a specified area and timely halts the policy execution to complete the task. In contrast, DoReMi~\cite{guo2023doremi} fails to stop the policy execution in time, leading to task failure. Red boxes indicate the occurrence of failures, while green boxes signify successful task execution.
   }
\label{fig: sup cliport demo 3}
\end{figure*}

\cref{fig: sup cliport demo 3} shows the ``Sweep Half the Blocks'' task, where our framework precisely counts the blocks within a specified area and timely halts the policy execution to complete the task. In contrast, DoReMi~\cite{guo2023doremi} fails to stop the policy execution in time, leading to task failure.

% \cref{fig: sup cliport demo 4} 展示了 ``Use Rope to Close the Opening Square'' 这个task，我们框架可以有效地发现开口方块被绳子闭合，并及时停止policy执行以成功完成任务，而DoReMi无法及时停止policy执行，导致虽然最后成功闭合，但是执行时间超时，导致最终任务失败。

\begin{figure*}[t]
\centering
\includegraphics[width=\linewidth]{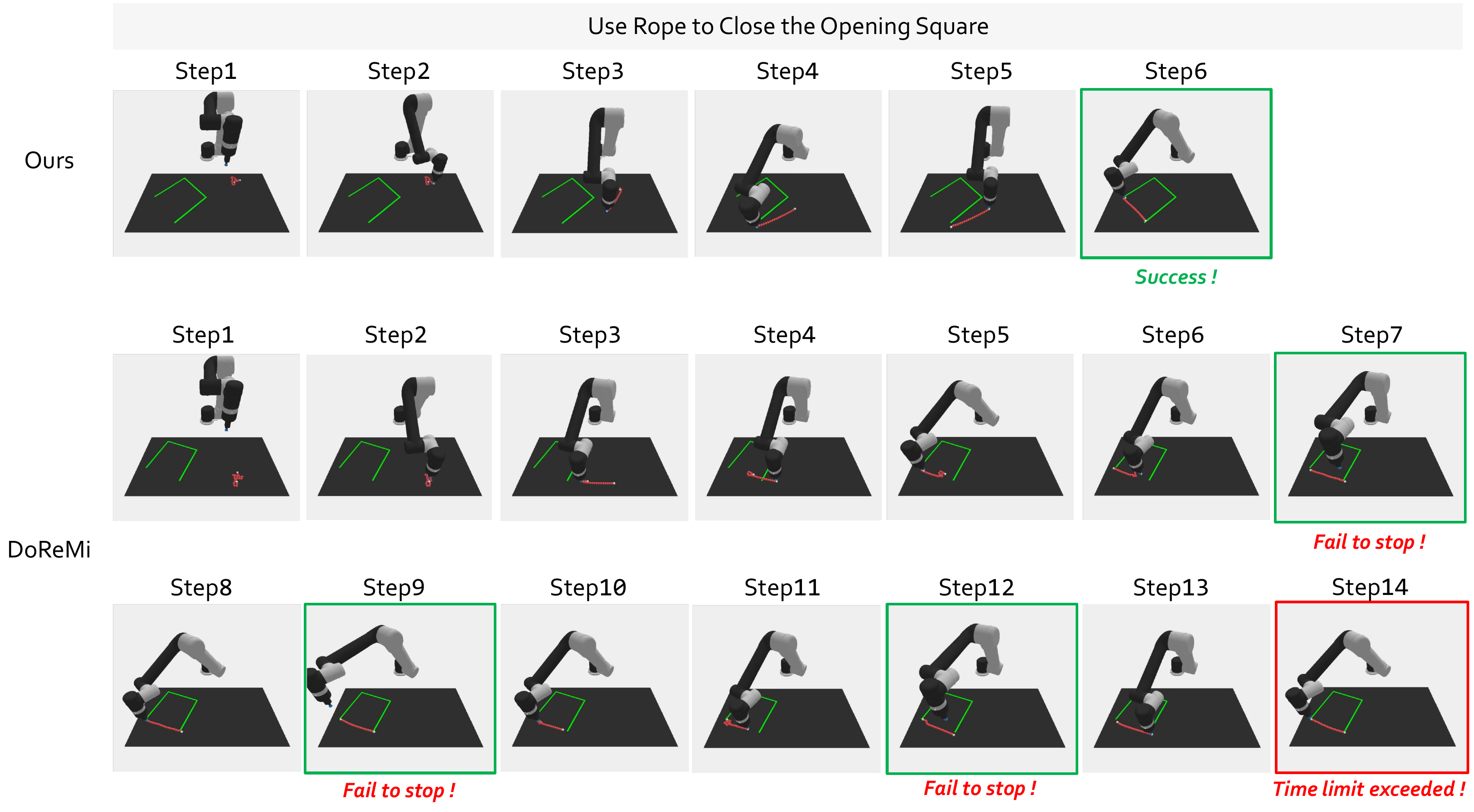}
   \caption{
   Demonstration of ``Use Rope to Close the Opening Square'' and comparison to baseline. We show that our framework effectively detects when the rope closes the opening square and promptly stops the policy execution to complete the task successfully. Conversely, DoReMi fails to halt the policy execution on time; although it eventually succeeds in closing the opening, the excessive execution time results in task failure. Red boxes indicate the occurrence of failures, while green boxes signify successful task execution.
   }
\label{fig: sup cliport demo 4}
\end{figure*}

\cref{fig: sup cliport demo 4} depicts the ``Use Rope to Close the Opening Square'' task. Our framework effectively detects when the rope closes the opening square and promptly stops the policy execution to complete the task successfully. Conversely, DoReMi fails to halt the policy execution on time; although it eventually succeeds in closing the opening, the excessive execution time results in task failure.

\begin{table}[t]
\caption{Following the Omnigibson evaluation protocol, we report the average success rate under disturbance (SR) and execution time to assess the impact of {\mname} and various failure detection modes on overall framework performance.}

% \vspace{-3mm}
\scriptsize
\centering
\setlength{\tabcolsep}{2.5pt}
\begin{tabular}{l|c|cc|cc|cc}
\bottomrule[1pt]
\multirow{2}{*}{ID} & \multirow{2}{*}{Method}        & \multicolumn{2}{c|}{Slot Pen} & \multicolumn{2}{c|}{Stow Book} & \multicolumn{2}{c}{Pour Tea}  \\
                        &                                & SR(\%) $\uparrow$ & Time(s)$\downarrow$ & SR(\%) $\uparrow$ & Time(s)$\downarrow$ & SR(\%) $\uparrow$ & Time(s)$\downarrow$\\
\hline
A                   &   LISA            &30.00      &126.89       &42.50      &118.93        &24.00      &218.92 \\
B                   &   PixelLM         &29.50      &134.10       &41.00      &124.26        &24.50      &214.04 \\
\hline
C                   &  Only Proactive   &37.50      &130.15       &50.00      &109.47        &32.50      &192.23 \\
D                   &  Only Reactive    &42.50      &157.63       &57.50      &147.95        &35.50      &284.15 \\
\hline
E                   &   \textbf{Ours}            &\textbf{47.50}      &\textbf{101.85}       &\textbf{65.00}      &\textbf{93.08}        &\textbf{40.00}      &\textbf{174.55} \\
%  ReKep~\cite{huang2024rekep}   & 30   & 20     & 10     &10          &  -                 &-           
%                                & 40   & 30     &30      &20          &  -                 &-            
%                                & 20   & 20   & 20      & 10          &  -                 &-           
             
%                                \\
% \hline
%  ~~+DRM   & 40   & 10     & 20     &20          & 177.84                & 54.54          
%                                & 50   & 40  & 20       &40          & 127.17                & 38.67   
%                                & 0   & 0   & 0     & 0          &  -                    & -           

%                                \\
 
%  ~~\textbf{+Ours}              & \textbf{60}   & \textbf{50}     & \textbf{40}     &\textbf{40}          &  \textbf{101.85}                 & \textbf{25.82}           
%                                & \textbf{70}   & \textbf{60}     &\textbf{70}      &\textbf{60}          &  \textbf{93.08}                  & \textbf{18.67}            
%                                & \textbf{50}   & \textbf{40}     & \textbf{40}     & \textbf{30}          & \textbf{174.55}                & \textbf{44.19}            

%                                \\

\bottomrule[1pt]
\end{tabular}
\label{tab: rebuttal omnigibson}
% \vspace{-4mm}
\end{table}

\subsection{OmniGibson}
\label{supsubsec: omniGibson demonstrations}

\begin{figure*}[t]
\centering
\includegraphics[width=\linewidth]{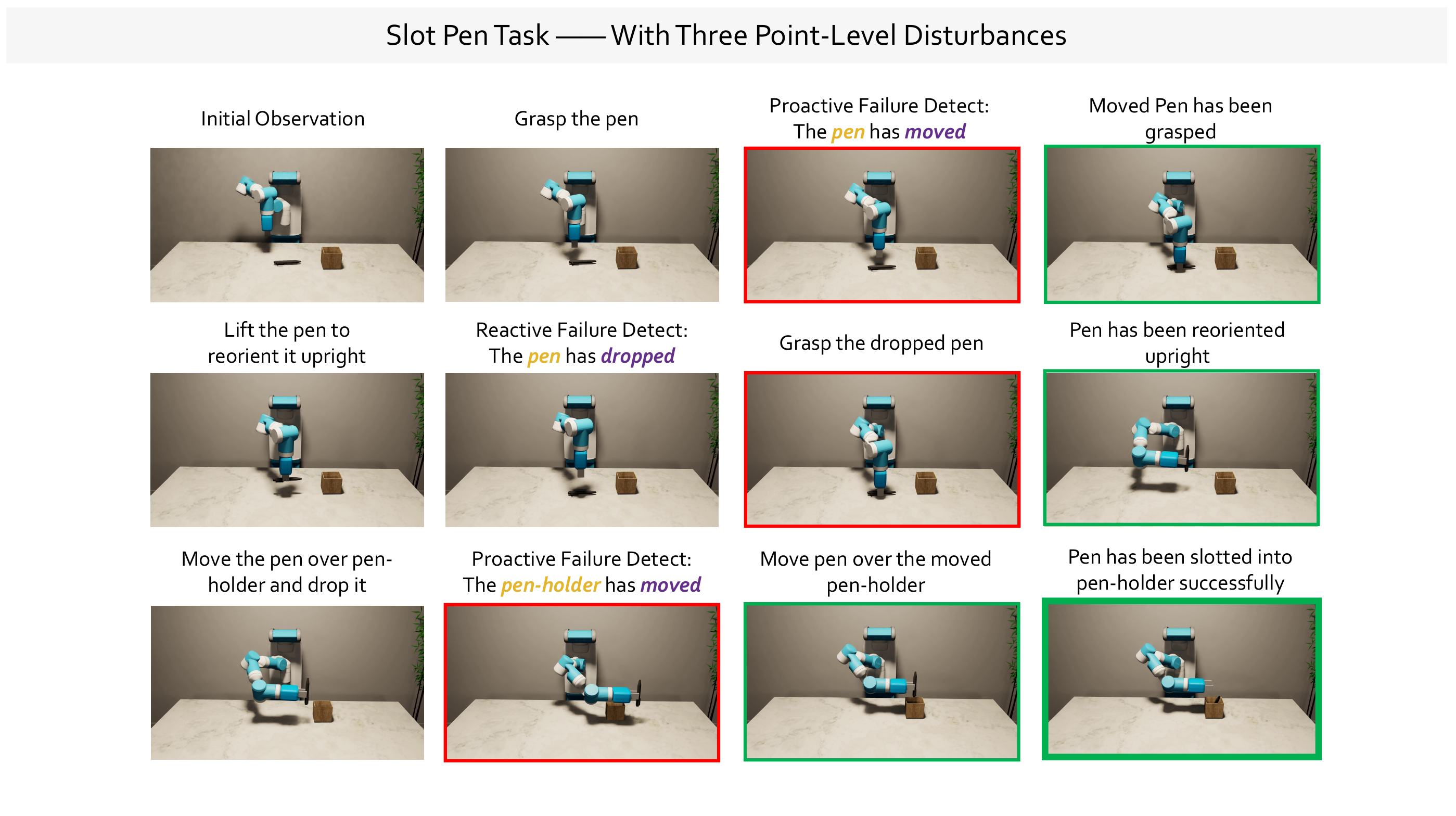}
   \caption{
   Demonstration of ``Slot Pen''. We show how our framework detects failures and assists in recovery when facing point-level disturbances. Red boxes indicate the occurrence of failures, light green indicates the recovery with subgoal success and dark green boxes signify successful task execution.
   }
\label{fig: sup omnigibson demo 1}
\end{figure*}

\begin{figure*}[t]
\centering
\includegraphics[width=\linewidth]{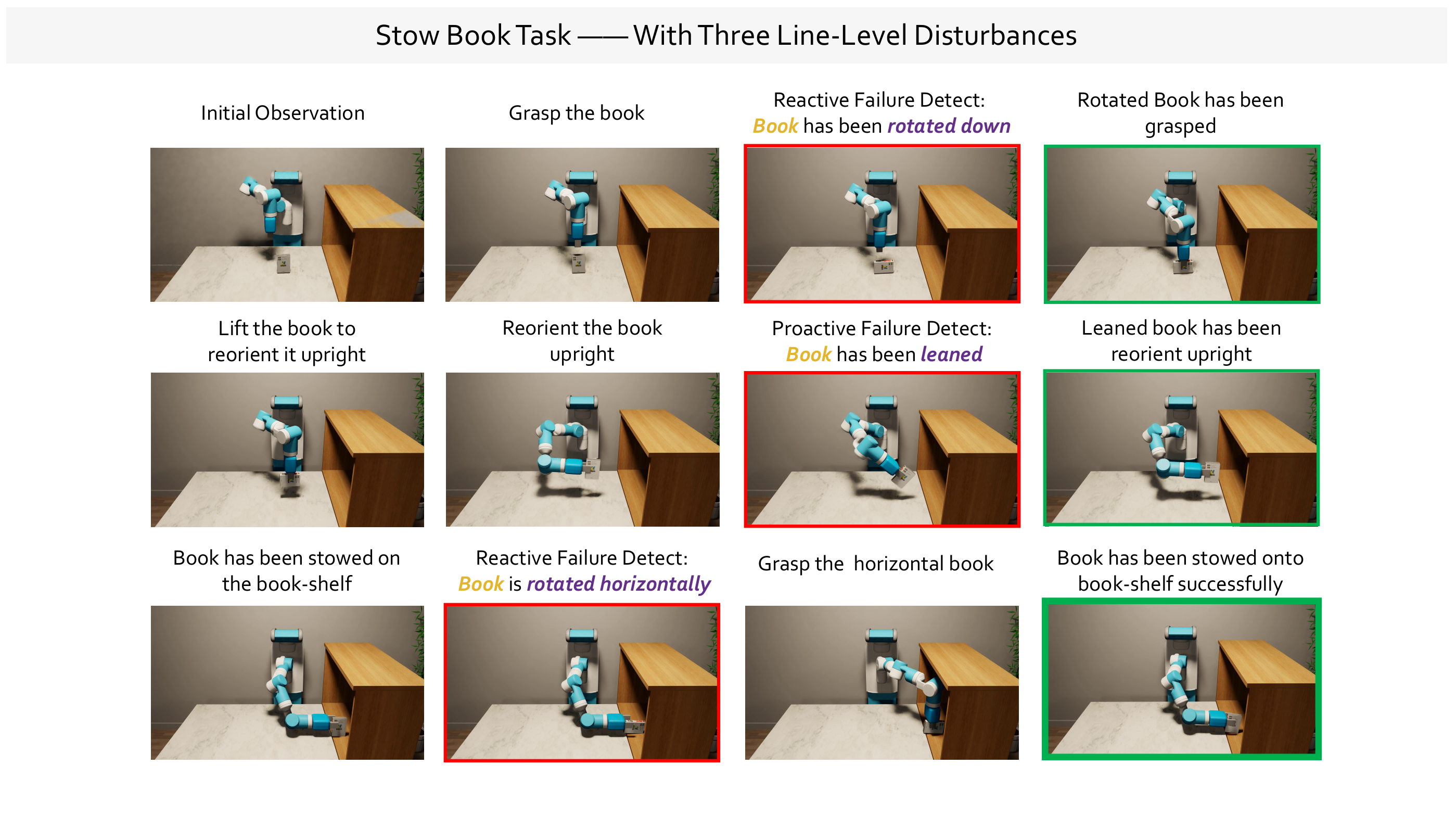}
   \caption{
   Demonstration of ``Stow Book''. We show how our framework detects failures and assists in recovery when facing line-level disturbances. Red boxes indicate the occurrence of failures, light green indicates the recovery with subgoal success and dark green boxes signify successful task execution.
   }
\label{fig: sup omnigibson demo 2}
\end{figure*}

\begin{figure*}[t]
\centering
\includegraphics[width=\linewidth]{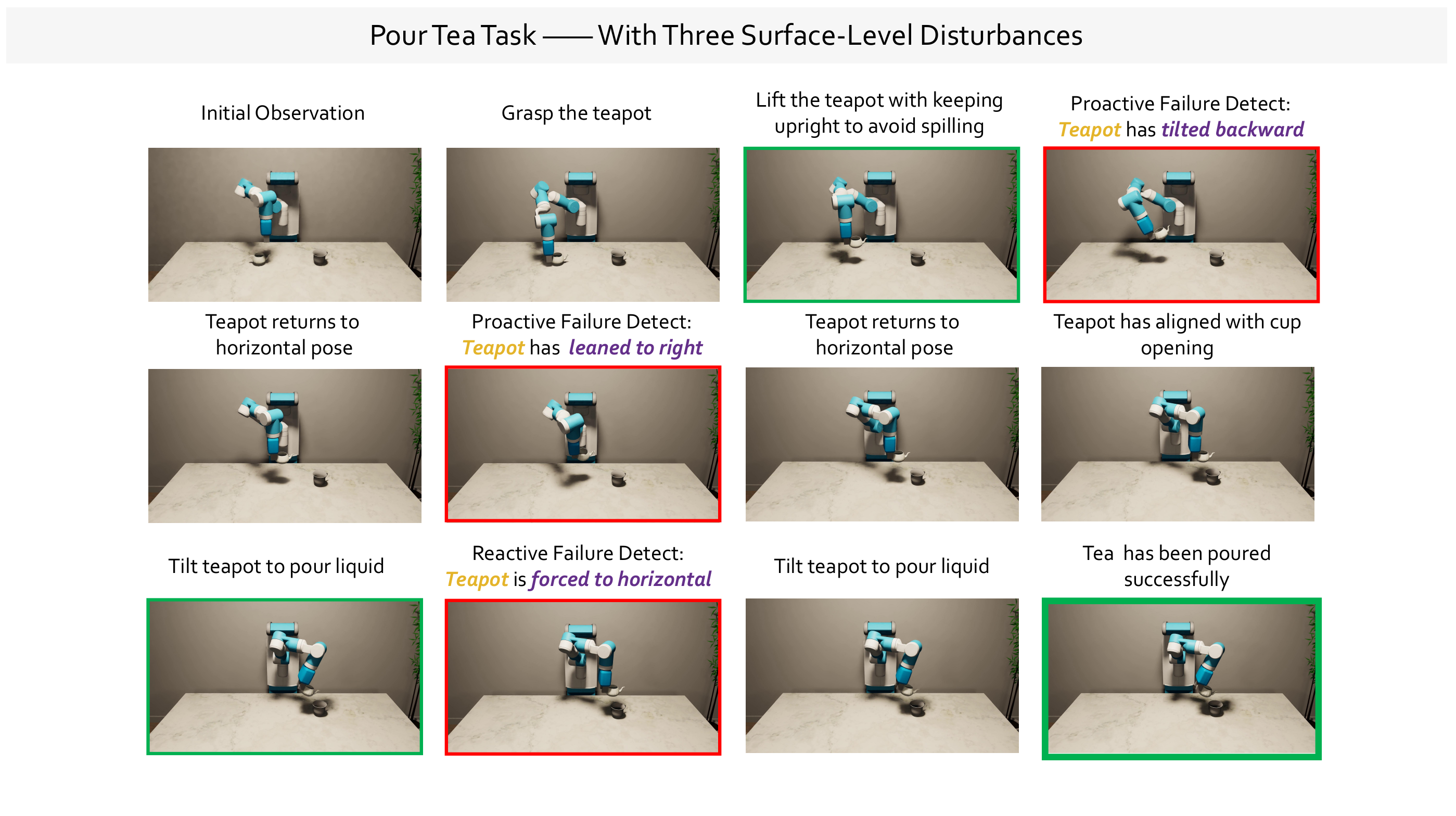}
   \caption{
   Demonstration of ``Pour Tea''. We show how our framework detects failures and assists in recovery when facing surface-level disturbances. Red boxes indicate the occurrence of failures, light green indicates the recovery with subgoal success and dark green boxes signify successful task execution.
   }
\label{fig: sup omnigibson demo 3}
\end{figure*}

As shown in \cref{fig: sup omnigibson demo 1}, \cref{fig: sup omnigibson demo 2} and \cref{fig: sup omnigibson demo 3}, we show how our framework detects failures and assists in recovery when facing point-, line- and surface -level disturbances.

\subsection{Real-world Evaluation}
\label{supsubsec: real-world demonstrations}

% \cref{fig: sup real demo} 展示了 ``Clear all objects on table except for animals.'' 这个task，我们框架实现了 reactive failure detection~(\eg, 在人类从机械臂手中remove物体后 detect unexpected failure)和proactive failure detection~(\eg, 在机械臂抓取的过程中发现抓取的目标物体移动从而prevent foreseeable failure)，有效提供了任务的成功率以及减少任务执行时间。

\begin{figure*}[t]
\centering
\includegraphics[width=\linewidth]{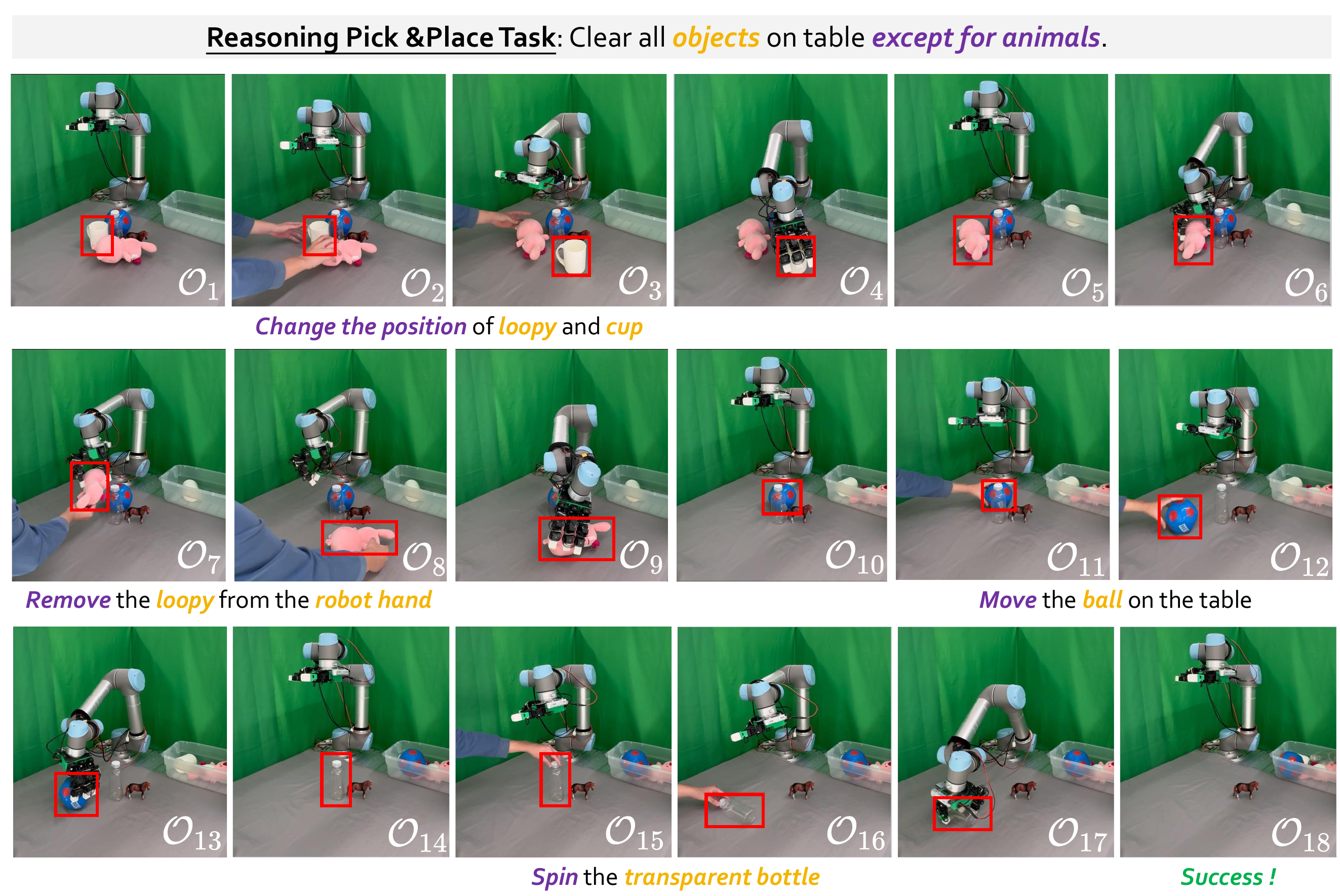}
   \caption{
   Demonstration of ``Clear all objects on the table except for animals''. We show that our framework achieves both reactive failure detection (\eg, detecting unexpected failures when humans remove objects from the robot's grasp) and proactive failure detection (\eg, identifying target object movement during grasping to prevent foreseeable failures). 
    This effectively enhances the task success rate and reduces the execution time.
   }
\label{fig: sup real demo}
\end{figure*}

\cref{fig: sup real demo} demonstrates the task ``Clear all objects on the table except for animals'', where our framework achieves both reactive failure detection (\eg, detecting unexpected failures when humans remove objects from the robot's grasp) and proactive failure detection (\eg, identifying target object movement during grasping to prevent foreseeable failures). 
This effectively enhances the task success rate and reduces the execution time.

\subsection{Constraint-aware segmentation}
\label{supp: constraint-aware segmentation demo}

% 这里我们展示了更多的有关 constraint-aware segmentation的结果，包括instance 和 part level的结果。为了展示泛化性，我们选用了out-or-distrubtion（OOD）的数据，包括 REFLECT的RoboFail Dataset, Open6DOF benchmark的 dataset, 以及 RT-1 dataset。同时我们也额外展示了Omnigibson仿真中的分割结果。

\begin{figure*}[t]
\centering
\includegraphics[width=\linewidth]{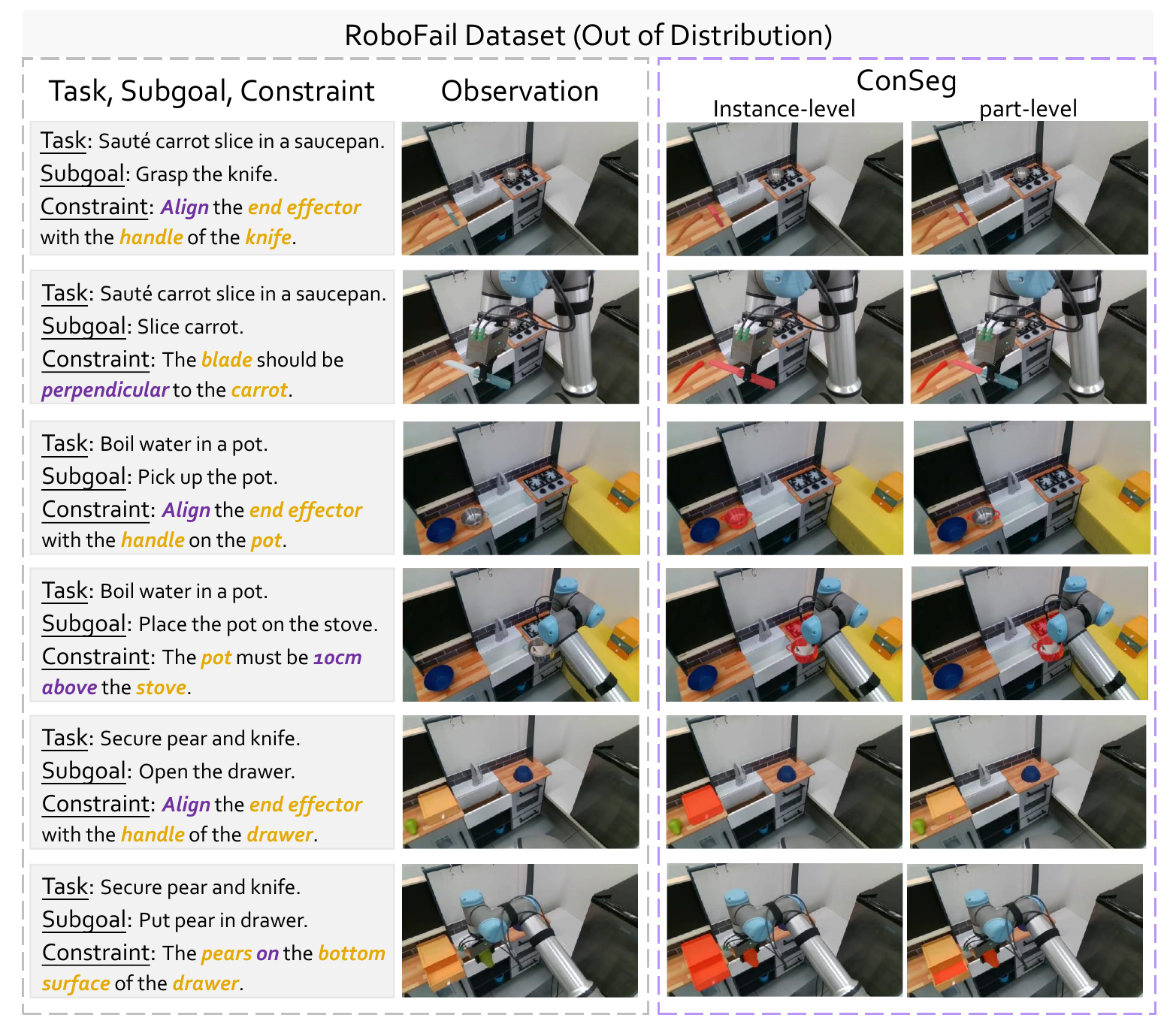}
   \caption{
   Visualization of constraint-aware segmentation for the RoboFail Dataset~\cite{liu2023reflect}. This dataset is not included in the training data.
   }
\label{fig: sup conseg robofail}
\end{figure*}

\begin{figure*}[t]
\centering
\includegraphics[width=0.9\linewidth]{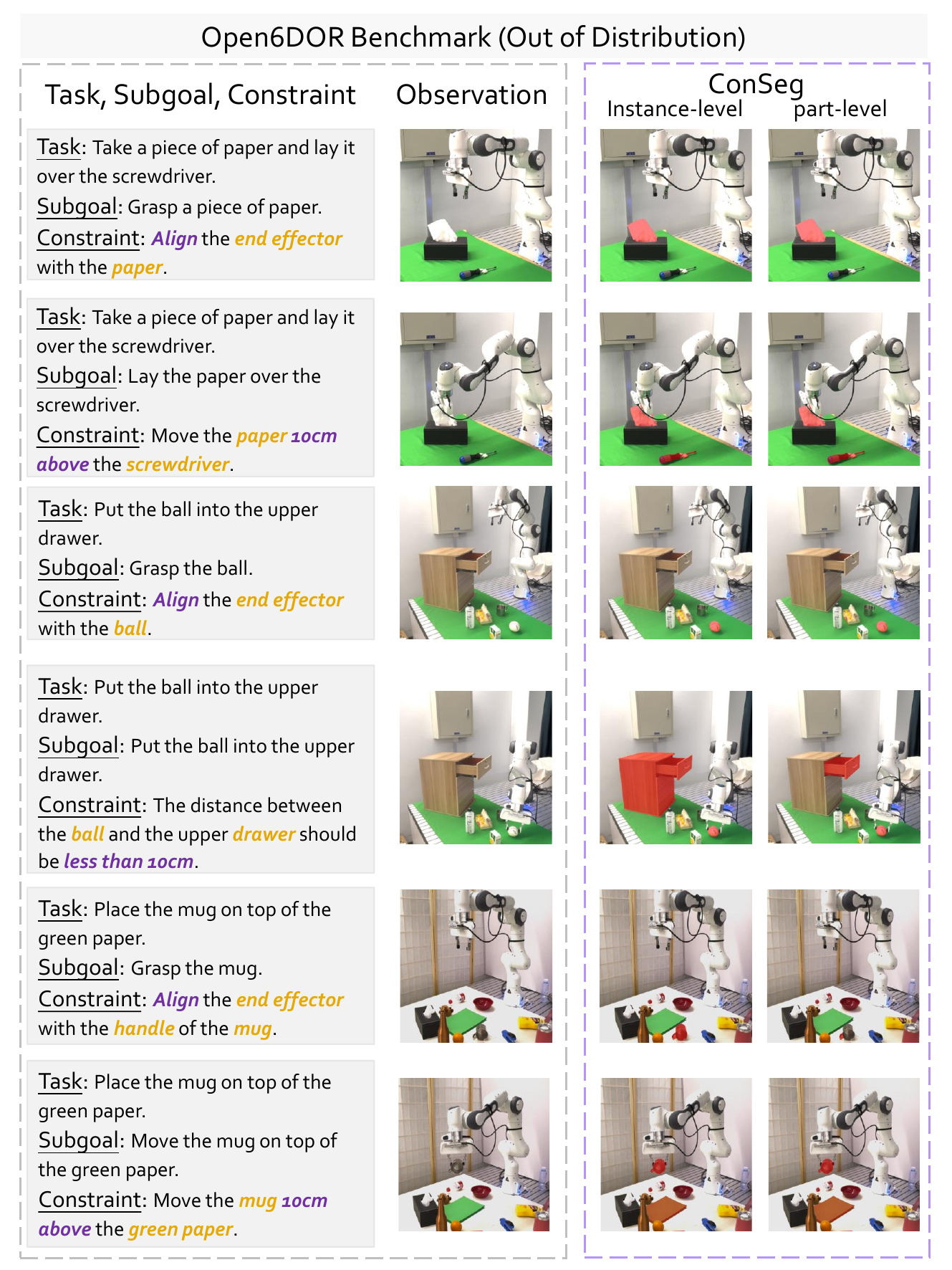}
   \caption{
   Visualization of constraint-aware segmentation for the Open6DOF~\cite{dingopen6dor}. This dataset is not included in the training data.
   }
\label{fig: sup conseg open6dor}
\end{figure*}

\begin{figure*}[t]
\centering
\includegraphics[width=\linewidth]{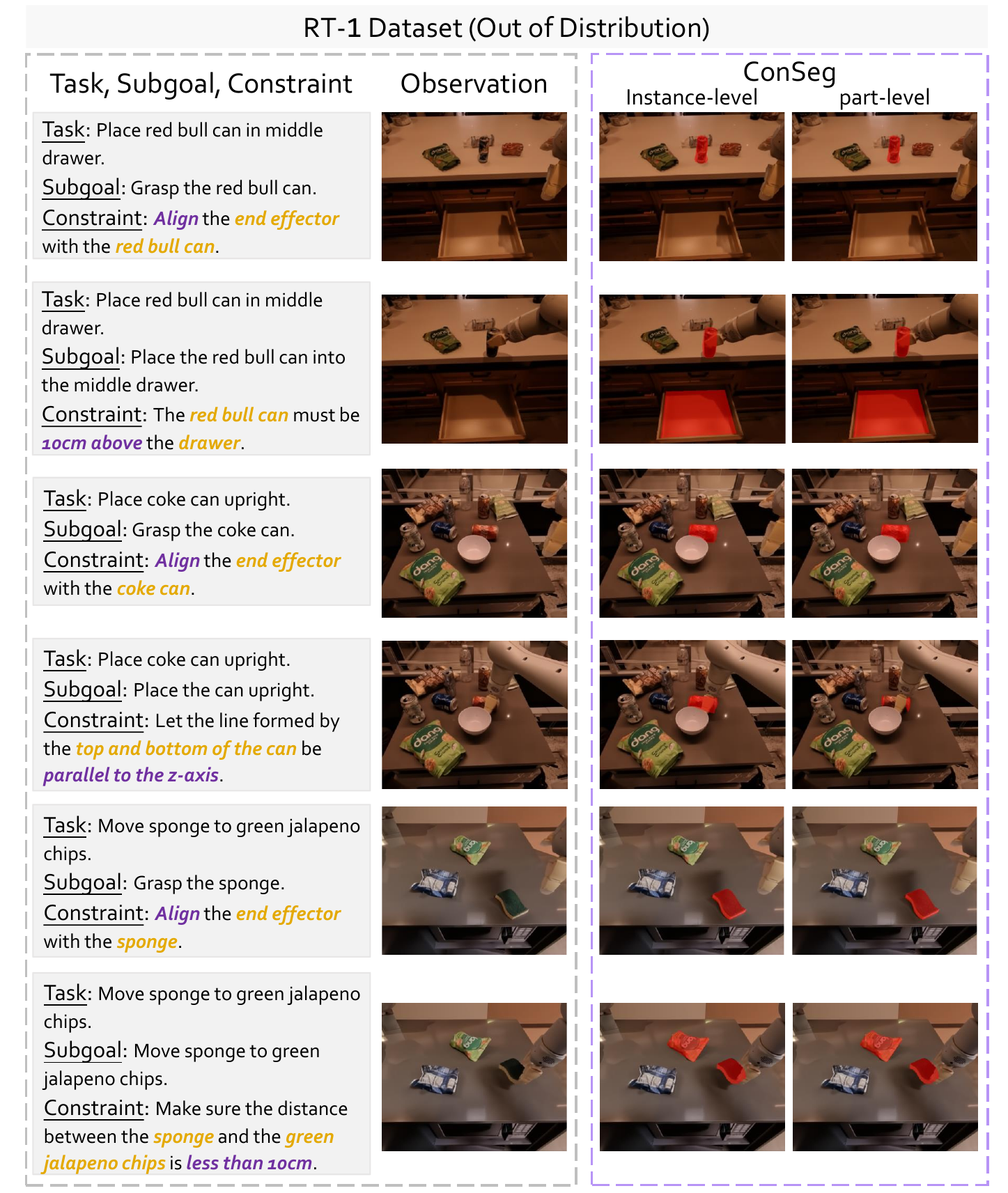}
   \caption{
   Visualization of constraint-aware segmentation for the RT-1 dataset~\cite{brohan2022rt}. This dataset is not included in the training data.
   }
\label{fig: sup conseg rt1}
\end{figure*}

\begin{figure*}[t]
\centering
\includegraphics[width=\linewidth]{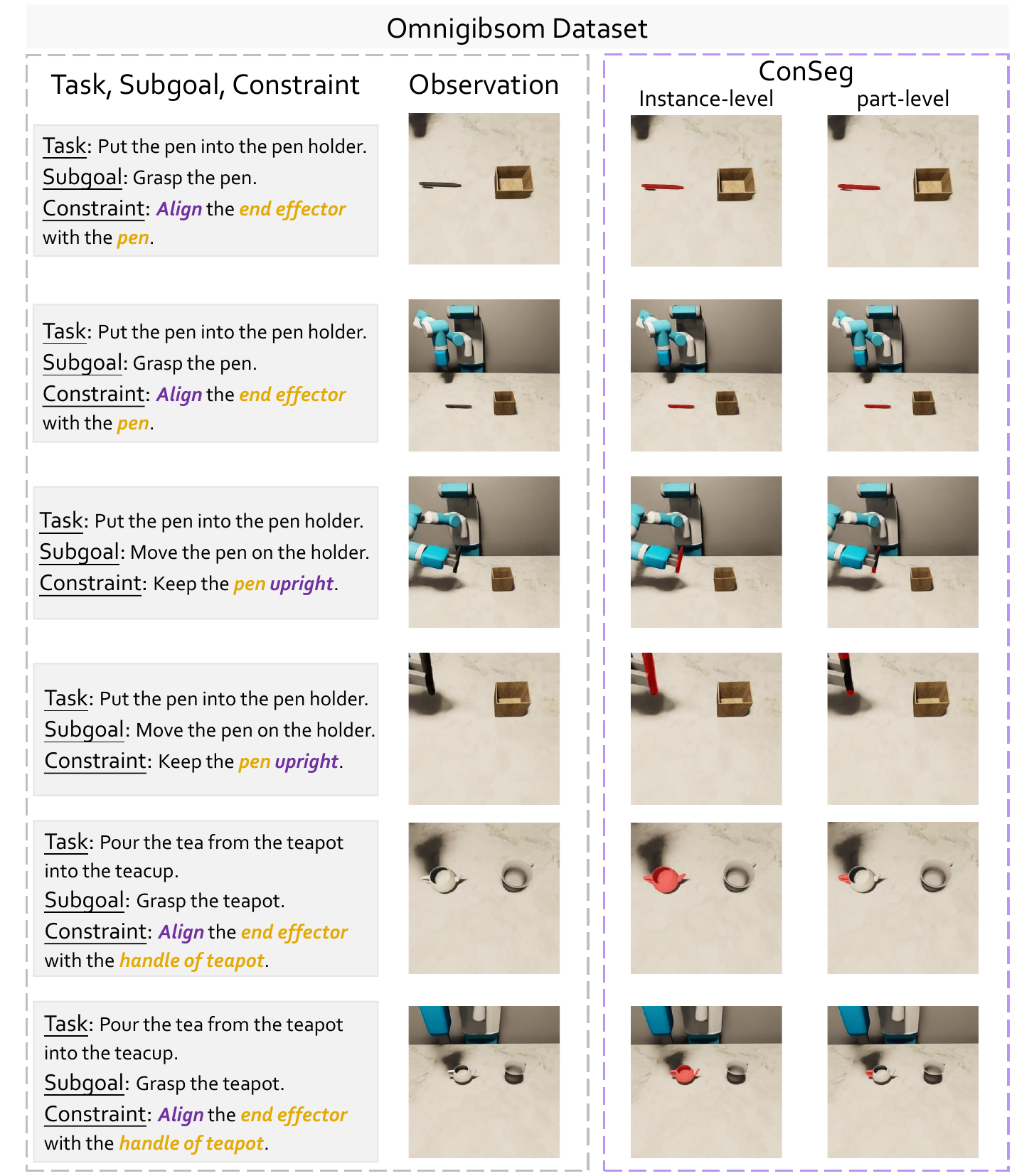}
   \caption{
   Visualization of constraint-aware segmentation for the Omnigibsom simulator. 
   % This dataset is not included in the training data.
   }
\label{fig: sup conseg omni}
\end{figure*}

As shown in \cref{fig: sup conseg robofail}, \cref{fig: sup conseg open6dor}, \cref{fig: sup conseg rt1}, and \cref{fig: sup conseg omni} we present additional results on constraint-aware segmentation, including instance and part-level results. 
\highlight{To demonstrate generalizability, we utilize out-of-distribution (OOD) data, including the RoboFail Dataset from REFLECT~\cite{liu2023reflect}, datasets from the Open6DOF benchmark~\cite{dingopen6dor}, and the RT-1 dataset~\cite{brohan2022rt}.} Additionally, we showcase segmentation results from the OmniGibson.

\subsection{Prompts}
\begin{figure*}[t]
\centering
\includegraphics[width=\linewidth]{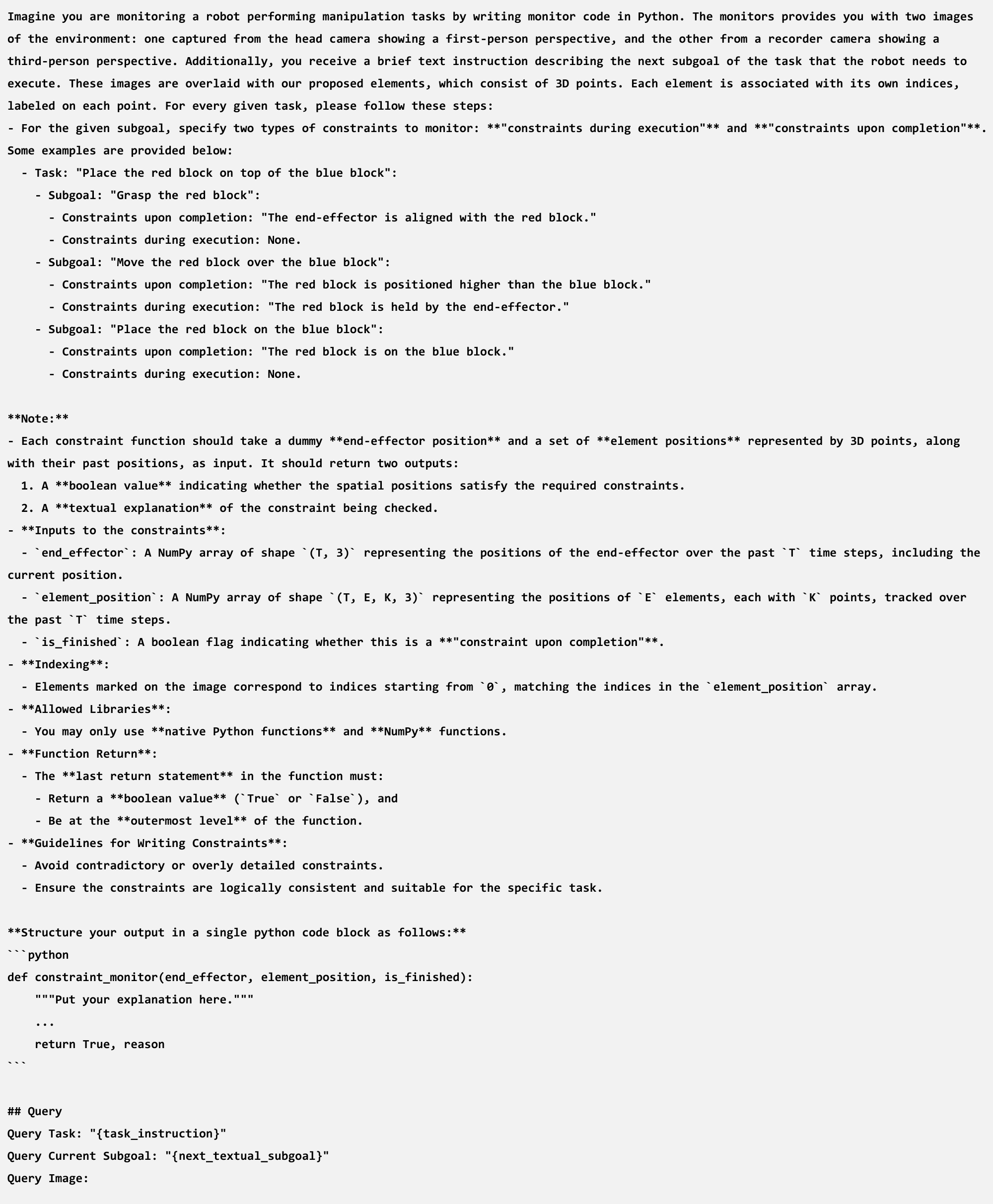}
   \caption{
   Prompt of monitor code generation. We use this prompt, combined with additional task instructions, the current subgoal, and images from two perspectives, to enable an off-the-shelf VLM, \ie, GPT-4o~\cite{achiam2023gpt}, to generate Python code for monitoring.
   }
\label{fig: sup prompt}
\end{figure*}

As illustrated in \cref{fig: sup prompt}, we detail the prompt used to invoke an off-the-shelf VLM, \ie, GPT-4o~\cite{achiam2023gpt}, to generate Python code for monitoring.

\newpage

\end{document}